\documentclass[10pt,twocolumn,letterpaper]{article}

\usepackage[pagenumbers]{cvpr} 
\usepackage{marvosym}

\usepackage{paralist}

\usepackage{xspace}
\usepackage{color}
\usepackage{multirow}
\usepackage{ifthen}

\long\def\ignorethis#1{}

\definecolor{gray}{rgb}{0.35,0.35,0.35}
\definecolor{MyBlue}{rgb}{0,0.2,0.8}
\definecolor{MyRed}{rgb}{0.8,0.2,0}
\definecolor{MyGreen}{rgb}{0.0,0.5,0.1}
\definecolor{MyGray}{rgb}{0.4,0.4,0.4}
\definecolor{purple}{RGB}{112,48,160}

\newlength\paramargin
\newlength\figmargin
\newlength\subfigmargin
\newlength\secmargin
\newlength\subsecmargin
\newlength\tabmargin
\newlength\eqmargin

\usepackage{array}
\newcolumntype{L}[1]{>{\raggedright\let\newline\\\arraybackslash\hspace{0pt}}m{#1}}
\newcolumntype{C}[1]{>{\centering\let\newline\\\arraybackslash\hspace{0pt}}m{#1}}
\newcolumntype{R}[1]{>{\raggedleft\let\newline\\\arraybackslash\hspace{0pt}}m{#1}}

\def\ie{i.e.,~}
\def\eg{e.g.,~}

\newcommand{\secref}[1]{Section~\ref{sec:#1}}

\newcommand{\figref}[1]{Fig.~\ref{fig:#1}}
\newcommand{\tabref}[1]{Table~\ref{tab:#1}}
\newcommand{\eqnref}[1]{Eq.\eqref{Eq:#1}}

\newcommand{\Paragraph}[1]{\noindent\textbf{#1}}

\definecolor{mycolor_blue}{RGB}{231,239,250}
\definecolor{mycolor_green}{RGB}{230,247,224}
\definecolor{mycolor_gray}{RGB}{236,236,236}
\definecolor{pearDark!20}{RGB}{212,230,241}

\definecolor{cvprblue}{rgb}{0.21,0.49,0.74}
\usepackage[pagebackref,breaklinks,colorlinks,allcolors=cvprblue]{hyperref}

\title{IC-Effect: Precise and Efficient Video Effects Editing via In-Context Learning}

\author{Yuanhang Li$^{1}$,\; Yiren Song$^{2}$,\; Junzhe Bai$^{1}$,\; Xinran Liang$^{1}$,\; Hu Yang$^{3}$,\; Libiao Jin$^{1}$,\; Qi Mao$^{1}$\textsuperscript{\Letter}\
\\ 
$^1$ School of Information and Communication Engineering, Communication University of China \\
$^2$ Show Lab, National University of Singapore  \quad $^3$ Baidu Inc., Beijing, China \\
 \href{https://cuc-mipg.github.io/IC-Effect/}{\textcolor{blue}{https://cuc-mipg.github.io/IC-Effect/}}
}

\begin{document}
\twocolumn[{
\renewcommand\twocolumn[1][]{#1}
\vspace{-0.3 cm}
\maketitle
\vspace{-0.8 cm}
\begin{center}
    \centering
    \includegraphics[width=1\linewidth]{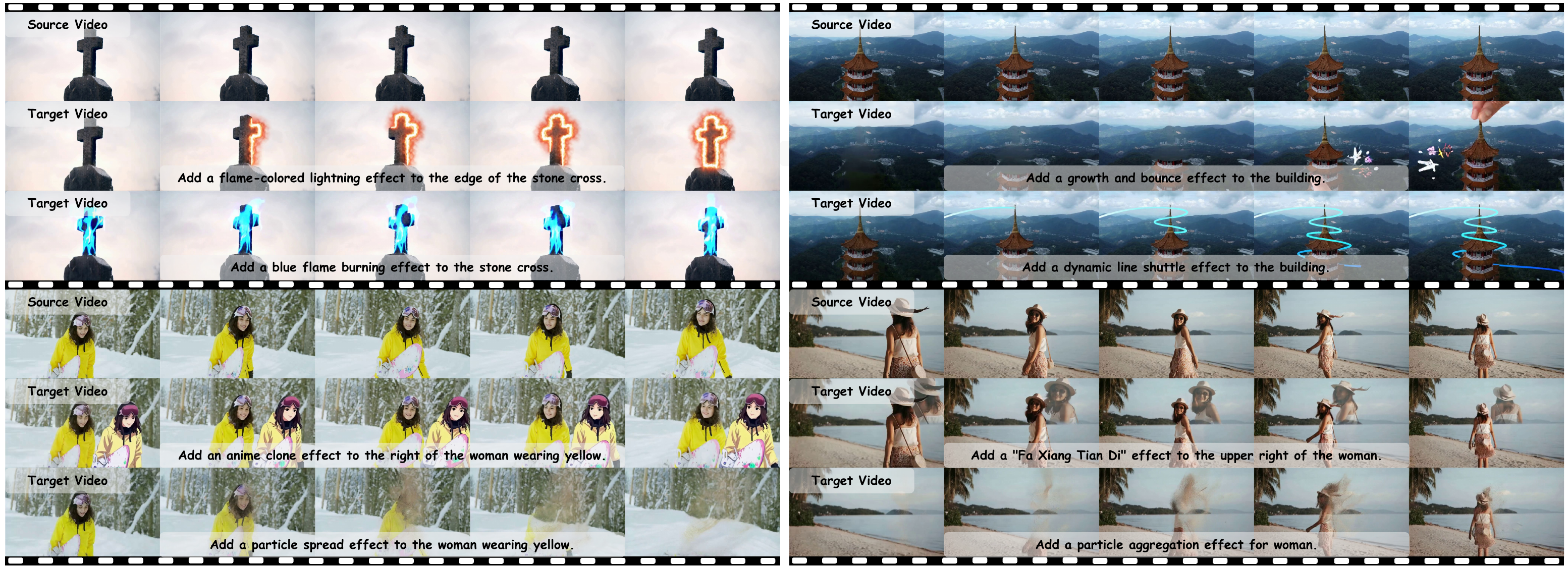}
    \vspace{-3 mm}
    \captionof{figure}{
    \textbf{Video VFX Editing Results of IC-Effect.} 
    Our IC-Effect enables precise video VFX editing aligned with textual instructions while preserving the complete spatiotemporal information of the source video. 
    The complete video is available in our supplementary materials.
    }
    \label{fig:teaser}
\end{center}
}]

\let\thefootnote\relax\footnotetext{\textsuperscript{\Letter} Corresponding Author: qimao@cuc.edu.cn}

\begin{abstract}

We propose \textbf{IC-Effect}, an instruction-guided, DiT-based framework for few-shot video VFX editing that synthesizes complex effects (\eg flames, particles and cartoon characters) while strictly preserving spatial and temporal consistency. 
Video VFX editing is highly challenging because injected effects must blend seamlessly with the background, the background must remain entirely unchanged, and effect patterns must be learned efficiently from limited paired data.
However, existing video editing models fail to satisfy these requirements. 
IC-Effect leverages the source video as clean contextual conditions, exploiting the contextual learning capability of DiT models to achieve precise background preservation and natural effect injection. 
A two-stage training strategy, consisting of general editing adaptation followed by effect-specific learning via Effect-LoRA, ensures strong instruction following and robust effect modeling.
To further improve efficiency, we introduce spatiotemporal sparse tokenization, enabling high fidelity with substantially reduced computation.
We also release a paired VFX editing dataset spanning $15$ high-quality visual styles. 
Extensive experiments show that IC-Effect delivers high-quality, controllable, and temporally consistent VFX editing, opening new possibilities for video creation.

\end{abstract}    
\section{Introduction}
\label{sec:intro}

Visual Effects (VFX) aim to create videos or edit existing ones by incorporating visually compelling elements such as flames, cartoon characters, or particle effects.
As a core technology in filmmaking, gaming, and virtual reality, VFX enrich visual storytelling, highlighting key elements, and creating immersive experiences. 
However, traditional video VFX workflows rely on complex animation design, computer-generated imagery (CGI), and professional post-production compositing. 
These processes incur high production costs, long turnaround times, and extensive manual intervention, which hinder personalized or real-time applications. 
Recent advances in text-to-video (T2V) generation~\cite{wan2025wan, yang2025cogvideox, kong2024hunyuanvideo} open new possibilities for automated VFX creation~\cite{liu2025vfx, mao2025omni}. 
However, video VFX editing, which automatically adds or modifies effects in an existing video, remains largely unexplored.

As a unique and higher-level video editing task, video VFX editing fundamentally differs from video VFX generation~\cite{liu2025vfx, mao2025omni}.
Its core objective is to seamlessly and realistically integrate visual effects into a source video while strictly preserving the spatial structure and temporal coherence of the original content. 
Although recent video editing models~\cite{jiang2025vace, decart2025lucyedit, cheng2024consistent, wu2025insvie, bian2025videopainter, gao2025lora} achieve significant progress across various editing tasks, they still struggle to meet the stringent requirements of video VFX editing.
Existing methods~\cite{jiang2025vace, decart2025lucyedit, cheng2024consistent, wu2025insvie} support global style changes or local content modification, but they often tolerate certain degrees of background or appearance changes, making it difficult to ensure pixel-level consistency with the source video. 
This limitation is unacceptable in video VFX editing, where the background must remain entirely unchanged. 
Mask-based editing methods~\cite{bian2025videopainter, gao2025lora} can preserve the unmasked regions, but they rely on pixel-accurate masks, which fundamentally conflicts with the goal of automated video VFX editing. 
Furthermore, unlike general video editing approaches that improve performance by leveraging large-scale data, producing high-quality paired VFX data is challenging, limiting the scalability of model training.
Effective video VFX editing must learn the unique patterns of effect injection from these high-quality paired samples to achieve physically consistent integration of effects into real scenes. 
These challenges make automated video VFX editing a largely unresolved problem.

To address above challenges, we propose \textbf{IC-Effect}, an instruction-guided, few-shot video VFX editing framework based on DiT-based T2V models.
IE-Effect leverages the contextual learning capability~\cite{huang2024context, chen2025edit, huang2025photodoodle, zhang2025easycontrol} of the DiT~\cite{peebles2023scalable} model, treating the source video as a clean contextual condition to provide the model with distortion-free information.
This enables the model to inject visual effects naturally while strictly preserving the original background content.
The framework adopts a two-stage training strategy: a pretrained DiT-based T2V model is first adapted into a universal video editor, then applying Effect-LoRA to extract effect-specific patterns a small set of paired VFX data for style customization while preserving the base model.
To improve efficiency, we introduce spatiotemporal sparse tokenization with position correction to preserve key features, reduce computation, and maintain accurate alignment between conditional tokens and generated frames for precise video VFX editing.

To mitigate data scarcity, we construct a dedicated dataset for video VFX editing that includes $15$ representative effect types—such as flames, anime clones, light particles, and bouncing.
Each sample contains a triplet annotation: the source video, the edited video with the target effect, and the corresponding textual description with spatiotemporal annotations.
All samples are carefully aligned in viewpoint, content, and motion to ensure reliable supervision for both training and evaluation.
To our knowledge, this is the first large-scale paired benchmark specifically designed for video VFX editing, filling a crucial gap in data resources and providing a reproducible platform for future research.

In summary, our contributions are as follows:

\begin{compactitem}
    \item We propose IC-Effect, the first instruction-guided video VFX editing framework built on DiT, achieving realistic effects while preserving background and temporal consistency.    

    \item We leverage DiT’s contextual learning with source video as clean condition, introducing spatiotemporal sparse tokenization and position correction for efficient and precise video VFX editing.

    \item  We construct the first high-quality paired video VFX dataset and demonstrating IC-Effect’s effectiveness and superiority through extensive qualitative and quantitative experiments.
    
\end{compactitem}
\section{Related Work}
\label{sec:related_work}

\subsection{Diffusion-Based Video Generation and Editing}

The success of diffusion models in image generation~\cite{rombach2022high, mao2024mag, hertz2022prompt, zhang2024ssr, zhang2024stable, zhang2025stable} drives their application to Video tasks~\cite{wu2023tune, guo2023animatediff, wang2023modelscope, chen2024videocrafter2, wan2025wan,li2024starvid, kong2024hunyuanvideo, yang2025cogvideox}.
Early approaches~\cite{wu2023tune, guo2023animatediff} extended image diffusion models by integrating temporal modules to produce short clips, while subsequent works~\cite{wang2023modelscope, chen2024videocrafter2} design specialized spatiotemporal architectures to enhance motion consistency. 
However, the U-Net backbone struggles with long-term dynamics. 
This limitation has been largely overcome by DiT-based models like Sora~\cite{videoworldsimulators2024}, CogVideoX~\cite{yang2025cogvideox}, HunyuanVideo~\cite{kong2024hunyuanvideo}, and Wan~\cite{wan2025wan}, which treat videos as unified spatiotemporal token sequences and employ 3D full attention to capture complex dependencies, significantly enhancing generation quality, smoothness, and semantic fidelity.

Concurrently, diffusion-based video editing models~\cite{geyer2023tokenflow, ku2024anyvv, liu2024video,jiang2025vace, ye2025unic, decart2025lucyedit, song2025layertracer, song2025makeanything, song2024processpainter} are also widely explored.
Initial strategies~\cite{geyer2023tokenflow, ku2024anyvv, liu2024video} inject source frames during inference but often suffer from flickering or artifacts. More stable alternatives~\cite{gu2024videoswap, hu2023videocontrolnet,  ma2024followpose, ma2024followyouremoji,  ma2025followyourclick, ma2025followyourmotion} encode structural priors—such as depth or optical flow—via auxiliary modules. 
Recent DiT-based image and video editing models~\cite{jiang2025vace, ye2025unic, decart2025lucyedit, huang2025arteditor, song2025omniconsistency} enable diverse editing by fusing source frames, reference conditions, and latent tokens at the token level. 
Yet, they fall short in video effects editing due to the abstract nature of visual effects and scarce paired training data. 
To bridge this gap, we propose a novel framework for text-guided video effects editing that synthesizes dynamic, semantically coherent effects while maintaining original content integrity and motion consistency.

\begin{figure*}[!t]
    \centering
    \includegraphics[width=\linewidth]{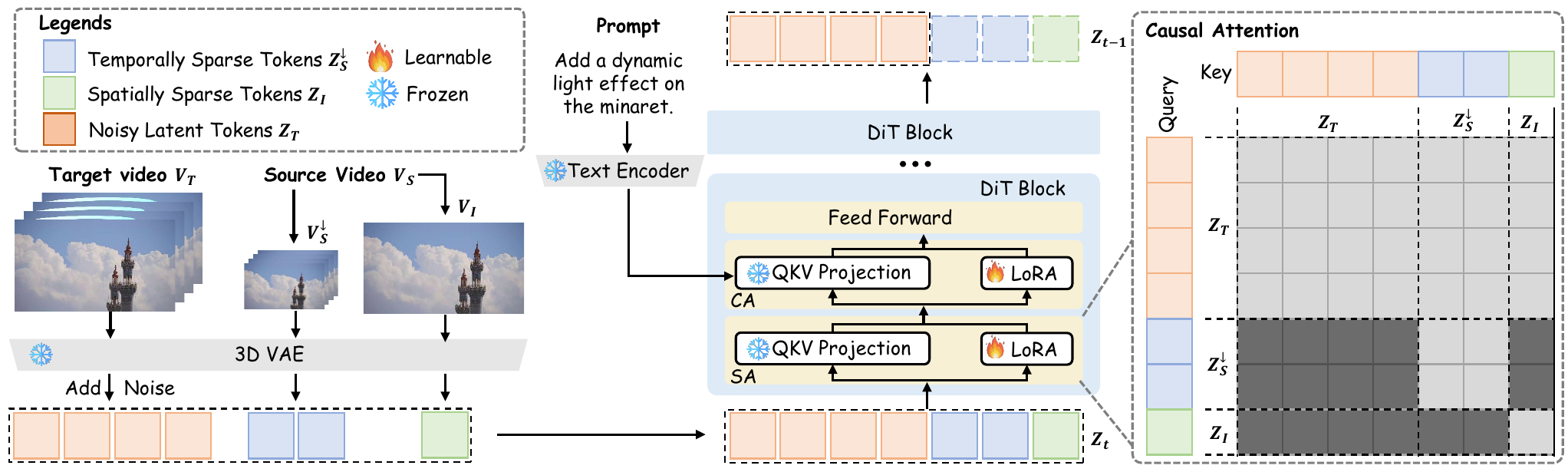}    
    \vspace{-3 mm}
    \caption{
    \textbf{The overall architecture and training paradigm of IC-Effect.}  
    Given a source video $V_S$, IC-Effect first tokenizes it into spatiotemporal sparse tokens $Z_S^{\downarrow }$ and $Z_I$. 
    These tokens are concatenated with noisy target tokens $Z_T$ along the token dimension to form a unified sequence, which is fed into a DiT module equipped with causal attention. 
    At the output, $Z_S^{\downarrow }$ and $Z_I$ are discarded, and only the target tokens $Z_T$ are decoded by a VAE to produce the edited video.
    During training, we first fine-tune the model with high-rank LoRA to acquire general video editing and instruction-following capabilities, and then further fine-tune it with low-rank LoRA on a small set of paired visual effects data to accurately capture the stylistic characteristics of diverse effects.       
    }
    \vspace{-3 mm}
    \label{fig:pipeline}
\end{figure*}

\subsection{Visual Effect Generation}

VFX creation aims to generate or edit images and videos to present hand-drawn, cartoon, or other creative artistic styles. 
In recent years, the development of artificial intelligence drives the partial automation of the VFX production pipeline~\cite{belova2021google, yu2023videodoodles}, significantly improving content creation efficiency. 
In the image domain, PhotoDoodle~\cite{huang2025photodoodle} first introduces a text-guided automatic image VFX editing method that adds semantically consistent visual elements to static images according to textual prompts. 
In the video domain, works such as OmniEffects~\cite{mao2025omni} and VFX-Creator~\cite{liu2025vfx} achieve text-guided video VFX generation. 
However, these methods mainly focus on synthesizing stylized video clips from scratch rather than performing precise and controllable dynamic VFX editing on existing videos. 
This paper aims to fill this gap by enabling creative visual style fusion based on textual descriptions, while preserving the structural integrity and motion coherence of the original video.

\subsection{Vision In-context Learning}

In-context learning (ICL) enables models to rapidly adapt to new tasks with only a few examples and is widely adopted in large language models~\cite{NEURIPS2020_1457c0d6}. 
Inspired by this, recent work explore the conditioning capability of DiTs, revealing their strong potential across various vision tasks~\cite{huang2024context, chen2025edit, huang2025photodoodle, zhang2025easycontrol, jiang2025vace}. 
Existing methods~\cite{huang2024context, chen2025edit, huang2025photodoodle, zhang2025easycontrol} typically introduce additional condition tokens, concatenate them with latent noise sequences, and jointly model them through 3D full attention to implicitly learn condition-content correlations. 
Recently, this framework is extended to the more challenging video generation domain~\cite{jiang2025vace, decart2025lucyedit, chen2025univid, ju2025fulldit}, achieving remarkable progress. 
For example, FullDiT~\cite{ju2025fulldit} achieves highly controllable text-to-video generation by integrating multimodal conditions like depth maps and camera trajectories.
However, since DiT’s computational cost grows quadratically with token count, directly concatenating long conditional token incurs substantial overhead, limiting the efficiency of high-resolution or long-duration video generation. 
In contrast, our method introduces spatiotemporally sparse condition tokens, fully exploiting DiT’s contextual understanding with minimal extra computation.

\section{Method}
\label{sec:method}

In this section, we introduce the overall architecture of IC-Effect in \secref{overall_architecture}. 
Next, we detail its key innovations: in-context conditioning for video VFX editing (\secref{ic_video}), Effect-LoRA for efficient effect learning (\secref{effect_LoRA}), and spatiotemporal sparse tokenization for reduced computational cost (\secref{STST}). 
Finally, we describe the proposed special-effects dataset in \secref{dataset}.

\subsection{Overall Architecture}
\label{sec:overall_architecture}

 The overall architecture of IC-Effect illustrated in \figref{pipeline}. We employ a two-stage training strategy:
 
\Paragraph{Pretraining Video-Editor.} We first pretrain a video editor on a large-scale video editing dataset.
This stage enables the model to accurately understand and respond to the text instruction, achieves controllable editing behavior, and establish a strong foundation for subsequent effect generation.

\Paragraph{Video Effect Fine-Tuning.} After pretraining, we introduce an Effect-LoRA module using Low-Rank Adaptation (LoRA)~\cite{hulora} and fine-tune it on a small set of paired video effect data. 
This stage focuses on learning the visual elements and stylistic characteristics of specific video effects, enabling precise modeling across different effect types. 
The proposed design supports efficient and lightweight customization, meeting the needs of diverse and personalized video effect generation.

\subsection{In-Context Conditioning for Video Editing}
\label{sec:ic_video}

Given a source video $V_S$ and its corresponding text instruction $T_E$, we aim to perform video VFX editing that aligns with the textual description while preserving the source video’s spatial layout and temporal coherence. 
Preserving the spatial layout and temporal coherence of the source video is essential for achieving faithful and artifact-free video VFX editing. 
To achieve this, we leverage the strong contextual modeling capability of the DiT-based T2V model and reformulate video VFX editing as a conditional generation problem, while minimizing modifications to the pretrained DiT architecture.

Specifically, we encode the source and edited video into latent representations $Z_S$ and $Z_T$ through a 3D VAE. 
Both representations share the same 3D rotary positional embeddings, which help the model capture stable relative spatiotemporal relationships during contextual modeling. 
$Z_S$ and $Z_T$ are patchified and concatenated into a unified token sequence, which is processed by the self-attention mechanism within the DiT blocks,
\begin{equation}
   \text{MMA}\left ( \left [ Z_T; Z_S \right ]  \right ) = \text{softmax} \left ( \frac{QK^{\top }}{\sqrt{d} }   \right ) V,
\label{Eq:mma}
\end{equation}
where $Z_T$ denotes the noisy latent tokens and $Z_S$ represents clean conditional tokens.
By using clean conditional tokens, the method preserves both the spatial structure and temporal motion information of the source video, providing faithful source information to $Z_T$ and preventing degradation in video quality during iterative denoising. 
Additionally, the attention mechanism allows the model to copy from the clean tokens or generate new content according to instructions, ensuring the edited video retains the source background information.
During training, we apply the flow matching loss~\cite{esser2024scaling} only to the latent tokens to guide the model in learning high-quality and structurally consistent video VFX editing.

\Paragraph{Causal Attention.}
Within the bidirectional attention of DiT, interactions between latent noise and  conditional tokens can cause clean conditional tokens to be fused with noisy representations, thereby degrading the quality of the generated results.
To prevent this, we introduce a causal attention mechanism.
In this mechanism, noise tokens attend to both themselves and clean conditional tokens, while conditional tokens attend only to themselves, avoiding any interaction with noise tokens. 
We realize this causal attention using a specifically designed attention mask:

\begin{equation}
M_{i,j} =
\begin{cases}
-\infty, & \text{if } i \notin Z_T \text{ and } j \in Z_T,\\
0, & \text{otherwise.}
\end{cases}
\label{Eq:mask}
\end{equation}

This causal attention effectively isolates clean conditional tokens from latent noise, preserving their fidelity and ensuring high-quality generation.

\subsection{Effect-LoRA}
\label{sec:effect_LoRA}

LoRA~\cite{hulora} enables efficient fine-tuning of large-scale pre-trained models by introducing trainable low-rank matrices while keeping the original weights frozen.
Given a pre-trained weight matrix $W_0 \in \mathcal{R}^{m \times n}$, LoRA introduces two low-rank matrices $A \in \mathcal{R}^{m \times r}$ $(r \ll m)$ and $B \in \mathcal{R}^{r \times n}$ $(r \ll n)$, where $r \ll \min(m,n)$, to model the parameter update:

\begin{equation}
   W = W_0 + AB.
   \label{Eq:lora}
\end{equation}

This formulation significantly reduces the number of trainable parameters while maintaining performance comparable to full-model fine-tuning.

To effectively learn the editing pattern of a specific video effect from limited paired VFX data, we propose Effect-LoRA. 
Effect-LoRA fine-tunes only a small number of trainable parameters, enabling efficient learning of video effect editing patterns while significantly reducing the risk of overfitting.
In our IC-Effect framework, the Video-Editor is trained on a large-scale paired dataset with a high-rank LoRA, which equips the model with strong instruction-following capability.
In contrast, Effect-LoRA adopts a low-rank LoRA specifically designed to capture the editing style of a single video effect. 
By guiding the behavior of the Video-Editor, Effect-LoRA enables the model to generate results that exhibit the desired effect style. 
When a new source video $V_S$ and a corresponding text instruction are provied, the model produces a target video $V_T$ that exhibits the visual effects described by the instruction.

\subsection{Spatiotemporal Sparse Tokenization}
\label{sec:STST}

The T2V model built upon the DiT~\cite{peebles2023scalable} architecture mainly relies on attention mechanisms, whose computational complexity is proportional to the square of the token length. 
Consequently, directly using the source video with the same resolution as the edited output as the contextual condition results in excessively long input tokens, thereby significantly increasing computational cost.
To overcome this limitation, we introduce a \textbf{spatiotemporal sparse tokenization} (STST) strategy that converts the source video $V_S$ into a set of spatiotemporally sparse tokens, effectively reducing the number of conditional tokens processed by the T2V model and improving inference efficiency.

Given a source video $V_S$, we do not directly convert it into latent tokens. Instead, we encode a downsampled version of the video along with its first frame into latent representations $Z_S^{\downarrow }$ and $Z_I$ using a 3D VAE. 
These representations are then patchified into temporally sparse tokens and spatially sparse tokens, respectively. 
Afterwards, $Z_S^{\downarrow }$, $Z_I$ and $Z_T$ are concatenated along the token dimension and fed into the DiT-based text-to-video (T2V) model. 
Formally, the transformation in \eqnref{mma} is expressed as follows:

\begin{equation}
   \text{MMA}\left ( \left [ Z_T; Z_S^{\downarrow }; Z_I \right ]  \right ) = \text{softmax} \left ( \frac{QK^{\top }}{\sqrt{d} }   \right ) V.
\label{Eq:mma_}
\end{equation}

To efficiently capture the spatiotemporal information from the source video, we sparsify it along both temporal and spatial dimensions. 
The temporally sparse tokens provide essential motion information, while the spatially sparse tokens supply complementary fine-grained spatial details. 
This spatiotemporal sparse tokenization achieves a balance between computational efficiency and visual fidelity, preserving the complete spatiotemporal characteristics of the source video. 
It provides reliable guidance for editing without incurring the high computational cost associated with tokenizing the full-resolution video.

Furthermore, \eqnref{mask} is reformulated as follows:
\begin{equation}
M_{i,j} =
\begin{cases}
0, & \text{if } i \in Z_T,\\
0, & \text{if } (i,j) \in Z_S^{\downarrow }  \text{ or }  (i,j) \in Z_I ,\\
-\infty, & \text{otherwise.}
\end{cases}
\label{Eq:mask_}
\end{equation}

\Paragraph{Spatiotemporal Position Correction.}
Although spatiotemporal sparse tokens significantly improve computational efficiency, they introduce a critical issue: spatiotemporal misalignment between the sparse conditional tokens and the target generation space.
This misalignment often leads to structural inconsistencies between the edited video and the source video, thereby degrading overall model performance.
To address this problem, inspired by OmniControl2~\cite{tan2025ominicontrol2}, we propose a spatiotemporal position correction technique.

For temporal sparse tokens $Z_S^{\downarrow }$, we establish explicit correspondences between the temporal sparse tokens and the target regions to prevent spatial misalignment:

\begin{equation}
   P_{Z_S^{\downarrow }}=P_{Z_T}\left ( n \cdot i, n \cdot j  \right ) .
\label{Eq:L_optim}
\end{equation}

For spatially sparse tokens $Z_I$, we use the positional encoding of the first frame of the noisy latent tokens $Z_T$ to avoid temporal misalignment. The spatiotemporal position correction technique is crucial, as it ensures that the spatiotemporally sparse tokens provide appropriate temporal and spatial information for video editing. 

\subsection{Video Effects Dataset}
\label{sec:dataset}

We construct the first VideoVFX dataset for text-guided video effect editing, covering $15$ high-quality effect categories with over $20$ curated videos.
The main categories include dynamic particle dispersion, particle dissipation and reassembly, line traversal, subject cartoon mirroring, flame combustion, graffiti, and architectural bouncing.
Each sample contains a source video $V_{S}$—from real-world scenes to portraits—and a target video $V_{T}$ with artist-designed effects exhibiting precise spatiotemporal coherence.
Effects include local stylization, animated overlays, motion-enhanced transitions, new dynamic elements, and structural modifications.
For each pair, we provide $V_{S}$, $V_{T}$, and a text instruction, enabling supervised learning for instruction-following video editing models.

\section{Experiment}
\label{sec:experiment}

\begin{figure*}[!t]
    \centering
    \includegraphics[width=\linewidth]{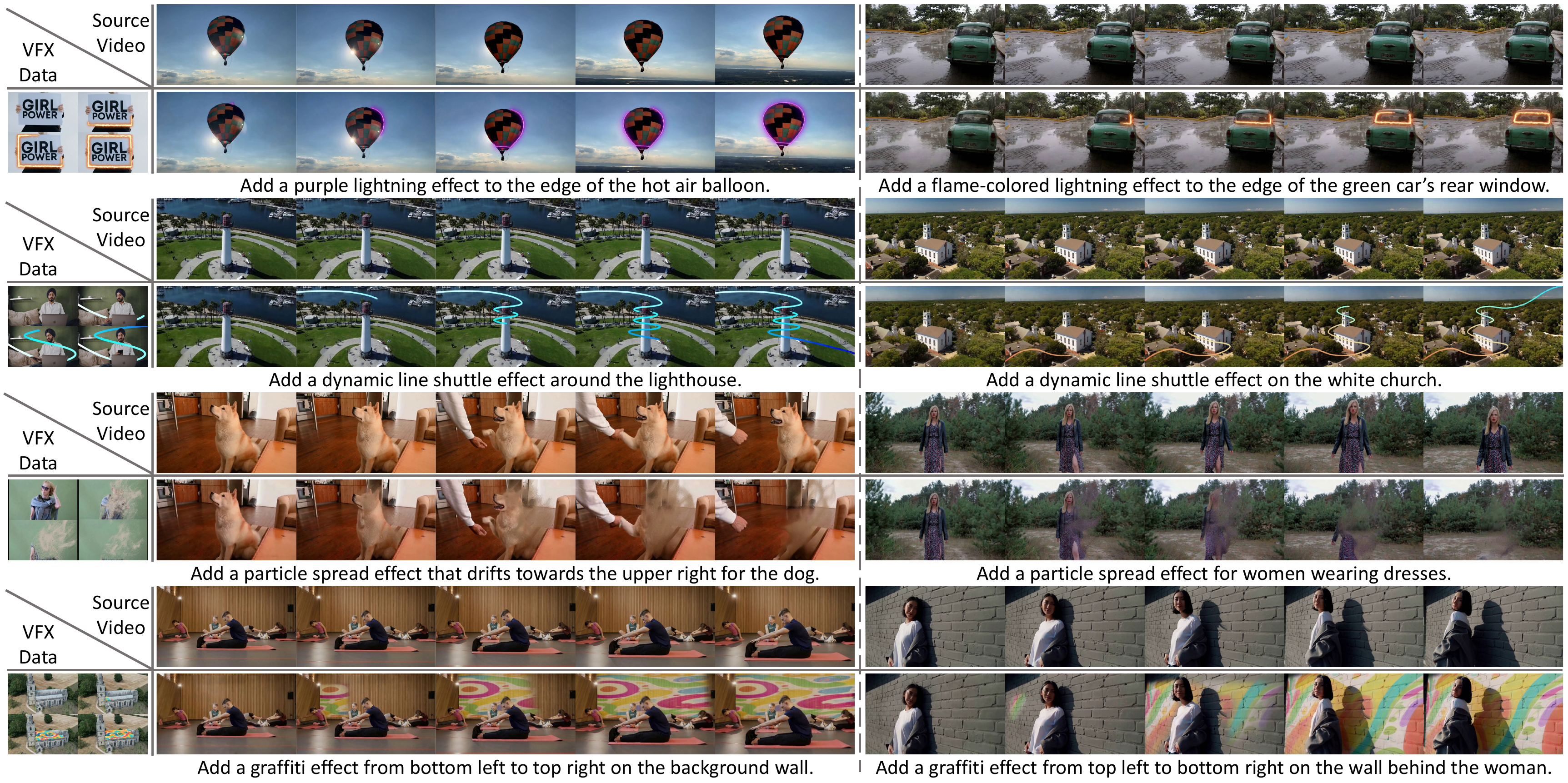}    
    \vspace{-5 mm}
    \caption{
    \textbf{Video VFX Editing Results of IC-Effect.} 
    IC-Effect accurately edits the source video following textual instructions, applying the visual effect styles present in the VFX data.
    The complete video is available in our supplementary materials.
    }
    \vspace{-3 mm}
    \label{fig:generate_result}
\end{figure*}

\subsection{Experimental Setup}

\Paragraph{Implementation Details.}
In the pretraining stage of Video-Editor, we initialize the DiT architecture with the parameters of Wan 2.2-A14B-T2V~\cite{wan2025wan} and train it on a self-constructed video editing dataset. 
All videos are resized to $224\times416$ with $81$ frames. 
The training uses four A800 GPUs with a batch size of $2$, a learning rate of $1\times10^{-4}$, and a LoRA rank of $96$ for $50,000$ optimization steps. 
The resulting LoRA weights are merged into the base T2V model to form Video-Editor, which serves as the backbone for subsequent tasks.
In the Effect-LoRA training stage, we further fine-tune Video-Editor on the paired video VFX editing dataset using two A800 HPUs for $1,000$ steps, with a LoRA rank of $32$, a batch size of $2$, and a learning rate of $1\times10^{-4}$. 
During inference, the model generates edited videos at a resolution of $480\times 832$ with $81$ frames.

\Paragraph{Baseline Methods.}
To evaluate the performance of our method, we compare it with several open-source approaches, including InsV2V~\cite{cheng2024consistent}, InsViE~\cite{wu2025insvie}, VACE~\cite{jiang2025vace}, and Lucy Edit~\cite{decart2025lucyedit}. 
For a fair comparison, we conduct experiments under two scenarios: common video editing and video VFX editing. 
In the Video VFX editing scenario, we fine-tune the attention layers of the above models using the same VFX dataset. 
Finally, we compare all the trained LoRA-based models with our IC-Effect framework.

\Paragraph{Benchmarks.}
We collect a total of $80$ video samples from the DAVIS~\cite{pont20172017} dataset and the Internet to evaluate the performance of the proposed Video-Editor. 
For the customized video effect editing task, we further construct a new benchmark dataset, entirely sourced from the Internet, consisting of 50 high-quality videos that cover both single-subject and multi-subject scenes. 
The subjects include humans, animals, buildings, and vehicles. This benchmark aims to comprehensively assess the model’s capability in customized video effect editing across diverse and realistic scenarios.

\Paragraph{Evalution Metrics.}
To evaluate the effectiveness of the proposed method, we adopt the following automatic evaluation metrics that analyze performance from multiple perspectives:
 1)Video Quality. We employ CLIP~\cite{radford2021learning} Image Similarity (CLIP-I) to measure temporal consistency by computing the cosine similarity between consecutive frames of the edited videos using the CLIP image encoder.
 2)Semantic Alignment. To assess the semantic consistency between the edited results and the text prompts, we use both the CLIP~\cite{radford2021learning} encoder and the ViCLIP~\cite{wang2024internvid} encoder to calculate frame-level and video-level similarities between the edited videos and the corresponding text prompts.
 3)Overall Quality. We adopt multiple sub-metrics from the VBench~\cite{huang2024vbench} toolkit, including Smoothness, Dynamic 
 Degree, and Aesthetic Quality, to comprehensively evaluate the visual fidelity and naturalness of the edited videos.
To further evaluate the structural preservation and effectiveness of the edited results, We further employ GPT-4o~\cite{gpt4o} to evaluate edited results along two dimensions: structural preservation, measuring spatial–temporal consistency with the source video, and effect accuracy, assessing alignment with the intended target (reference effect or textual prompt).

\subsection{Video VFX Editing Results}

\figref{generate_result} presents the results of IC-Effect in video VFX editing. 
IC-Effect accurately injects visual effects into the source video according to the given instruction while maintaining consistency between the generated effect styles and those in the training data, demonstrating excellent instruction-following ability. 
This performance benefits from training on high-quality paired data, which enables the model to learn robust effect generation and background preservation capabilities. 
When fine-tuned with Effect-LoRA on a limited dataset of paired effect data, IC-Effect continues to generate the specified effects stably and strictly preserves the spatiotemporal consistency of non-effect regions, avoiding color drift and structural distortion. 
Notably, IC-Effect maintains stable performance and a high success rate under different settings, exhibiting strong generalization and controllability, and achieves consistent editing results in production environments without selective sampling.

\begin{table*}[!t]
\centering
\caption{\textbf{Quantitative Comparison of Common Video Editing and Video VFX Editing.}
The best and second-best values are highlighted in \colorbox{pearDark!20}{blue} and \colorbox{mycolor_green}{green}, respectively.
}
\vspace{-1 mm}
\resizebox{\linewidth}{!}{%
\begin{tabular}{cccccccccc}
\hline
\multicolumn{2}{c}{} & \multicolumn{1}{c}{\textbf{Video Quality}}  & \multicolumn{2}{c}{\textbf{Semantic Alignment}} & \multicolumn{3}{c}{\textbf{Overall Quality}} & \multicolumn{2}{c}{\textbf{GPT Score}} \\ \cmidrule(r){3-3} \cmidrule(r){4-5} \cmidrule(r){6-8} \cmidrule(r){9-10}
\multicolumn{2}{c}{\multirow{-2}{*}{\textbf{Method/Metrics}}} & \textbf{CLIP-I} \textbf{($\uparrow$)} & \textbf{CLIP-T} \textbf{($\uparrow$)} & \textbf{ViCLIP-T} \textbf{($\uparrow$)} & \textbf{Smoothness} \textbf{($\uparrow$)} & \textbf{Dynamic Degree} \textbf{($\uparrow$)} & \textbf{Aesthetic Quality} \textbf{($\uparrow$)} & \textbf{Structural Preservation} \textbf{($\uparrow$)} & \textbf{Effect Accuracy} \textbf{($\uparrow$)}  \\ 
\midrule
&  InsV2V~\cite{cheng2024consistent} & \colorbox{mycolor_green}{0.9750} &  22.9317 &  \colorbox{mycolor_green}{21.7245} &  0.9798  &  \colorbox{mycolor_green}{0.6147} &  0.4723 &  3.9006  & 2.4452   \\
& InsViE~\cite{wu2025insvie}  &  0.9686 &  23.0331 &  20.5973 &  0.9695  &  0.5856 &  0.4877 &  3.6468  & 2.3687  \\
& VACE~\cite{jiang2025vace}  &  0.9698 &  \colorbox{mycolor_green}{23.4918} &  21.4685 &  0.9803  &  0.4820 &  \colorbox{mycolor_green}{0.5190} &  3.8562  & 2.6218 \\
& Lucy Edit~\cite{decart2025lucyedit}  &  0.9721 &  22.8130 &  21.1405 &   \colorbox{mycolor_green}{0.9811}  &  0.5737 &  0.5071 &  \colorbox{mycolor_green}{4.0529} & \colorbox{mycolor_green}{3.1218} \\
\multicolumn{1}{c}{\multirow{-5}{*}{\rotatebox{90}{\textbf{Common}}}} & \textbf{Ours} &  \colorbox{pearDark!20}{0.9795} &   \colorbox{pearDark!20}{25.6774} &  \colorbox{pearDark!20}{24.3290}  &  \colorbox{pearDark!20}{0.9815} & \colorbox{pearDark!20}{0.6374} &  \colorbox{pearDark!20}{0.5261} &  \colorbox{pearDark!20}{4.3824}  & \colorbox{pearDark!20}{4.2652}   \\
\midrule
&  InsV2V~\cite{cheng2024consistent} &  0.9767 &  25.6705 &  23.8159 &  0.9882  & 0.2019 &  \colorbox{mycolor_green}{0.5612} &  3.9326  & 1.9807   \\
&  InsViE~\cite{wu2025insvie}  & 0.9770 &  25.6530 &  23.7950 &  0.9858  & 0.3258 &  0.5158 &  4.1012  & 2.8988  \\
& VACE~\cite{jiang2025vace}  &  0.9782 &  25.5283 &  24.5594 &  \colorbox{mycolor_green}{0.9897} & 0.2788 &  0.5401 &  4.2105  & 3.6988 \\
& Lucy Edit~\cite{decart2025lucyedit}  &  \colorbox{mycolor_green}{0.9784} &  \colorbox{mycolor_green}{27.0618} &  \colorbox{mycolor_green}{26.2143}  & 0.9894 & \colorbox{mycolor_green}{0.3653} &  0.5546 &  \colorbox{mycolor_green}{4.3461}  & \colorbox{mycolor_green}{3.9423} \\
\multicolumn{1}{c}{\multirow{-5}{*}{\rotatebox{90}{\textbf{VFX}}}} & \textbf{Ours} &  \colorbox{pearDark!20}{0.9786} &  \colorbox{pearDark!20}{27.2321} &  \colorbox{pearDark!20}{26.6312}  & \colorbox{pearDark!20}{0.9911} & \colorbox{pearDark!20}{0.3771} &  \colorbox{pearDark!20}{0.5823} &  \colorbox{pearDark!20}{4.7947}  & \colorbox{pearDark!20}{4.5614}   \\
\bottomrule
\end{tabular}%
}
\label{tab:tab_baseline}
\end{table*}

\begin{figure*}[!t]
    \centering
    \includegraphics[width=\linewidth]{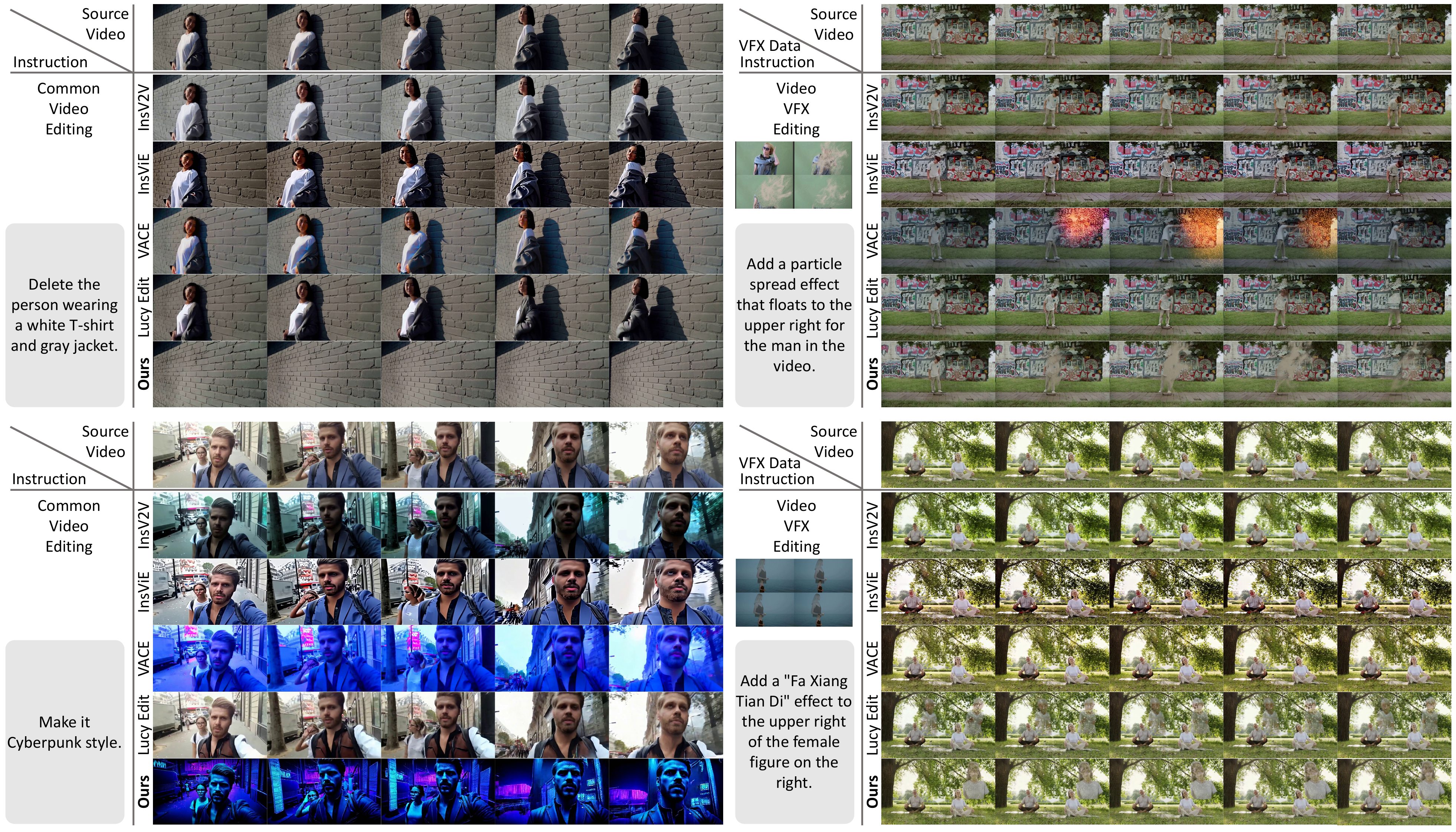}    
    \vspace{-5 mm}
    \caption{
    \textbf{Qualitative Comparison of Video Editing.} 
    IC-Effect demonstrates strong performance in instruction following, spatiotemporal consistency, and editing quality.
    The complete video is available in our supplementary materials.
    }
    \vspace{-3 mm}
    \label{fig:qualitative}
\end{figure*}

\subsection{Comparison with Baselines}

\Paragraph{Qualitative Comparison.}
\figref{qualitative} present qualitative comparisons of the proposed method on general video editing and customized video effect editing tasks. 
As shown in \figref{qualitative}, compared with state-of-the-art video editing methods, Video-Editor demonstrates stronger instruction-following capability, accurately performing the modifications described in the textual prompts while effectively preserving the structural and content consistency of non-edited regions.
This advantage mainly stems from the use of our carefully constructed high-quality training dataset and the task-oriented model design.
In the customized video VFX editing task (\figref{qualitative}), our method significantly outperforms the baselines. 
The generated videos exhibit higher visual quality, with added effects that align closely with the original ones in terms of style and motion dynamics, faithfully adhering to the textual instructions while minimizing content drift and unintended modifications to non-target areas. 
These results demonstrate that our method achieves a well-balanced trade-off between visual coherence and precise, controllable editing, substantially enhancing the reliability and naturalness of complex video effect editing.

\Paragraph{Quantitative Comparison.}
As shown in \tabref{tab_baseline}, our method outperforms all baseline approaches across all evaluation metrics on the general video editing task, achieving the best overall objective performance. 
This demonstrates that Video-Editor possesses stronger comprehensive capabilities in terms of temporal consistency, semantic alignment, and overall visual quality.
For the video effect editing task, our method also achieves the best metrics. 
Notably, GPT-4o significantly outperforms baseline methods in both structure preservation and effect accuracy.
The significant improvements in these metrics further validate the superiority of the proposed method in editing precision and semantic consistency, highlighting its enhanced controllability and generation quality in complex, customized video editing scenarios.
%

\begin{table*}[!t]
\centering
\caption{\textbf{Quantitative Comparison of Ablation Studies.}
The best and second-best values are highlighted in \colorbox{pearDark!20}{blue} and \colorbox{mycolor_green}{green}, respectively.
}
\vspace{-1 mm}
\resizebox{\linewidth}{!}{
\begin{tabular}{ccccccccccc}
\hline
\multicolumn{1}{c}{} & \multicolumn{1}{c}{\textbf{Video Quality}}  & \multicolumn{2}{c}{\textbf{Semantic Alignment}} & \multicolumn{3}{c}{\textbf{Overall Quality}} & \multicolumn{2}{c}{\textbf{GPT Score}} & \multicolumn{2}{c}{\textbf{Inference overhead}} \\ \cmidrule(r){2-2} \cmidrule(r){3-4} \cmidrule(r){5-7} \cmidrule(r){8-9} \cmidrule(r){10-11}
\multicolumn{1}{c}{\multirow{-2}{*}{\textbf{Method/Metrics}}} & \textbf{CLIP-I} \textbf{($\uparrow$)} & \textbf{CLIP-T} \textbf{($\uparrow$)} & \textbf{ViCLIP-T} \textbf{($\uparrow$)} & \textbf{Smoothness} \textbf{($\uparrow$)} & \textbf{Dynamic Degree} \textbf{($\uparrow$)} & \textbf{Aesthetic Quality} \textbf{($\uparrow$)} & \textbf{Structural preservation} \textbf{($\uparrow$)} & \textbf{Effect accuracy} \textbf{($\uparrow$)} & \textbf{Time} & \textbf{GPU memory} \\ 
\midrule
  W/o STST & \colorbox{pearDark!20}{0.9792} &\colorbox{pearDark!20}{ 27.5514} &  \colorbox{mycolor_green}{26.3071} &   0.9885 & \colorbox{mycolor_green}{0.3806} &  \colorbox{pearDark!20}{0.5940} & \colorbox{pearDark!20}{4.8375}  & \colorbox{pearDark!20}{4.5975} & 5880s & 74.71GB  \\
 w/o $Z_I$  &  0.9367 &  20.9034 &  21.4638 &  0.9657 & \colorbox{pearDark!20}{0.8163}  &   0.3805 &  4.0123  & 3.6140 & 2617s & 63.74GB  \\
 w/o Pretrain  &  0.9224 &  26.7977 &  25.8940 &   \colorbox{mycolor_green}{0.9886} &  0.3762  &  0.5395 &  4.2471  &3.9134 & 2790s & 64.07GB \\
 w/o Effect-LoRA & 0.9749 &  25.0413 &  23.4875 &   0.9802 & 0.3709  & 0.5395 &  4.1428  & 3.9423 & 2790s & 64.07GB  \\
  \textbf{Ours} &  \colorbox{mycolor_green}{0.9786} &  \colorbox{mycolor_green}{27.2321} &  \colorbox{pearDark!20}{26.6312}  & \colorbox{pearDark!20}{0.9911} & 0.3771 &  \colorbox{mycolor_green}{0.5823} &  \colorbox{mycolor_green}{4.7947}  & \colorbox{mycolor_green}{4.5614} & 2790s & 64.07GB   \\
\bottomrule
\end{tabular}
}
\vspace{-3 mm}
\label{tab:tab_ablation}
\end{table*}

\subsection{User Study}

\begin{figure}[!t]
    \centering
    \includegraphics[width=\linewidth]{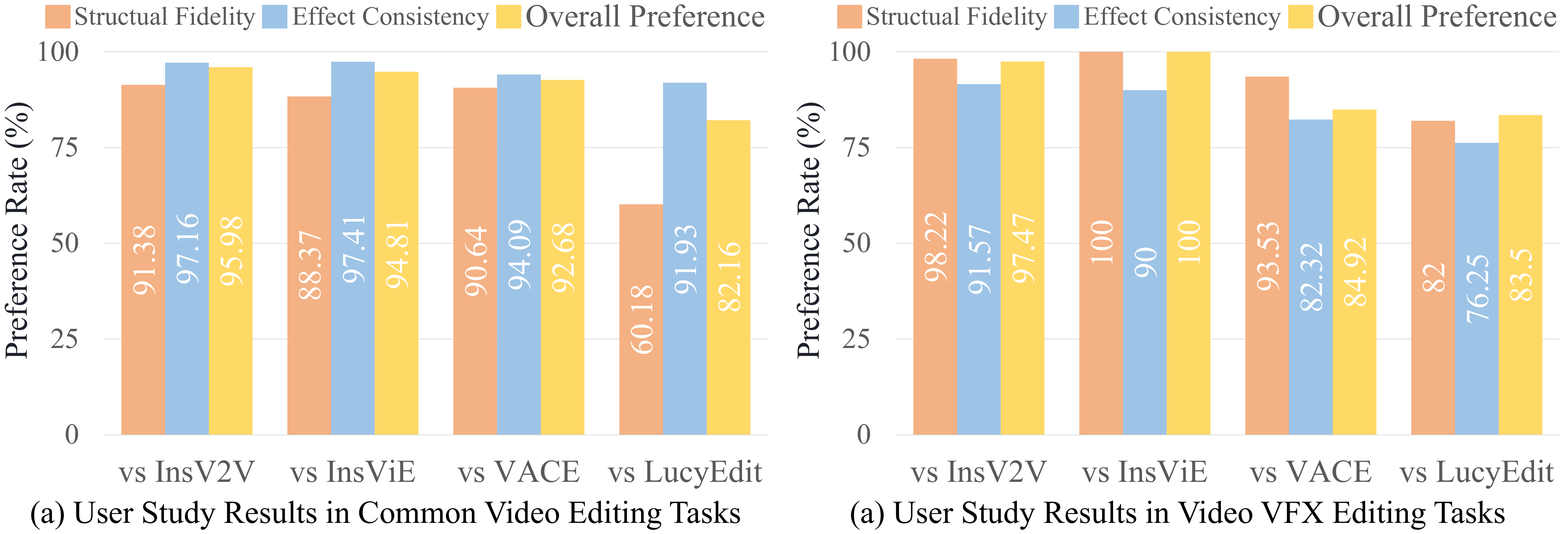}    
    \vspace{-5 mm}
    \caption{
    \textbf{Human Preference Study.} Our method is significantly more preferred
by users compared with the comparative methods.
    }
    \vspace{-3 mm}
    \label{fig:user_study}
\end{figure}

To further validate the effectiveness of the proposed method, we conduct a user study through an online questionnaire to evaluate user preferences in two scenarios: conventional video editing and customized video effect editing. 
Each participant is presented with a text instruction, a source video, and two edited results respectively generated by our method and a baseline. 
The two results are shown in a randomized order to prevent participants from inferring their sources and to minimize subjective bias.
The study follows an A/B testing protocol, where participants evaluate the results from three aspects: 
(1) instruction following, (2)  consistency between the edited video and the source video (\ie structual fidelity) and (3) Overall preference. 
A total of 20 participants are recruited for the study. 
As shown in \figref{user_study}, the statistical results demonstrate that, compared with baseline methods, our approach achieves higher user preference in both instruction adherence and source-video fidelity, verifying its effectiveness and advantage in real-world applications.

\subsection{Ablation Studies}

\begin{figure}[!t]
    \centering
    \includegraphics[width=\linewidth]{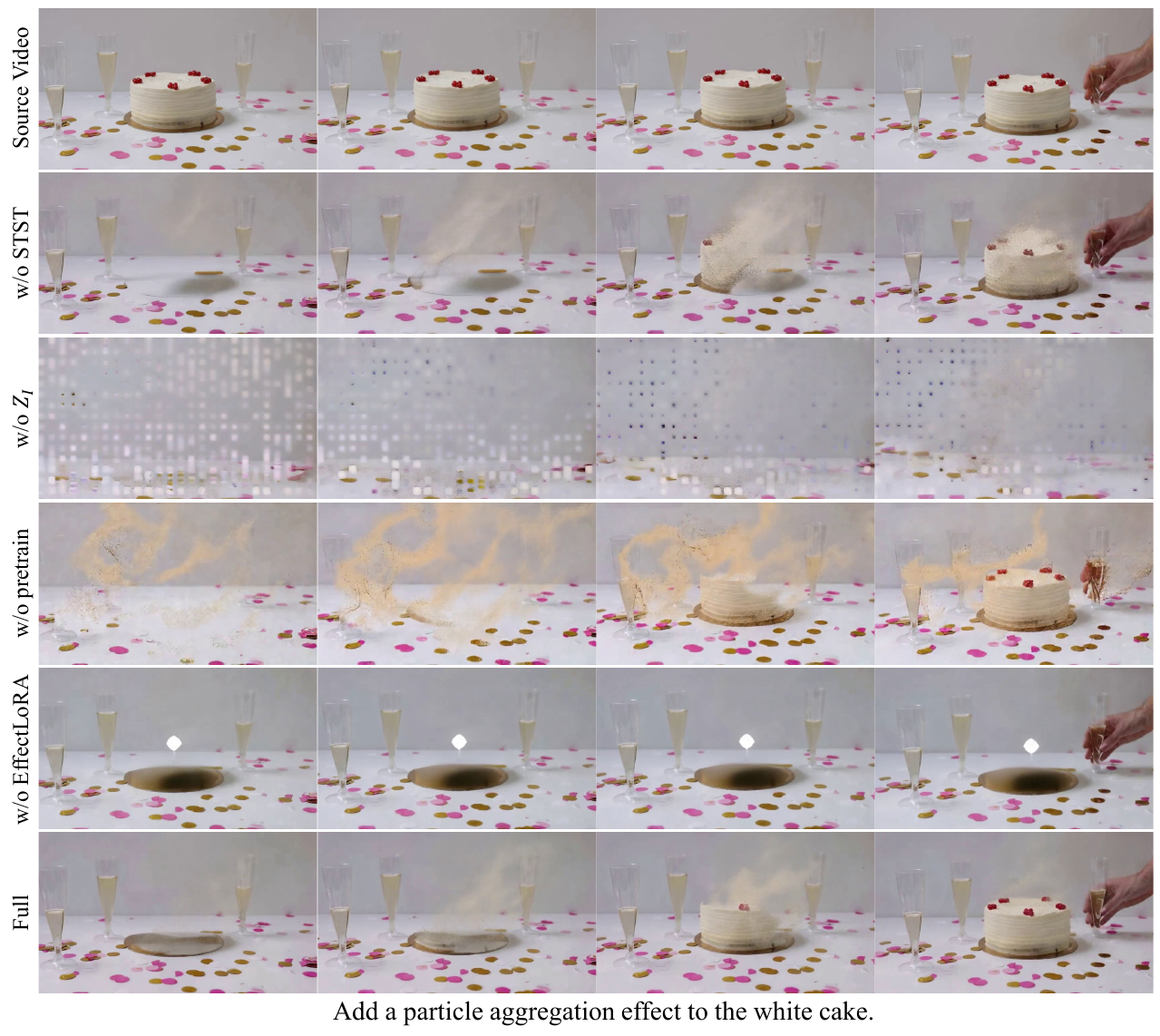}    
    \vspace{-5 mm}
    \caption{
    \textbf{Ablation Study of the Proposed Approach.} 
    }
    \vspace{-3 mm}
    \label{fig:ablation}
\end{figure}

To evaluate the effectiveness of the proposed strategies and modules, we conduct detailed ablation experiments, including the spatiotemporal sparse tokenization strategy, Video-Editor pretraining, and Effect-LoRA.
As shown in the \figref{ablation}, our method maintains editing quality comparable to the fully tokenized model, while various metrics in \tabref{tab_ablation} also approach those of the fully tokenized model, and the computational cost is significantly reduced.
Without using a high-quality first frame as the conditional input, relying solely on temporally sparse tokens fails to provide sufficient spatial details, resulting in noticeable blotchy artifacts and blurring, which is also reflected by the lowest metrics in \tabref{tab_ablation}.
In contrast, our spatiotemporal sparse tokenization strategy introduces minimal additional computation while substantially improving visual fidelity and spatiotemporal consistency, as also confirmed by the qualitative results in \tabref{tab_ablation}. 
Furthermore, skipping the pretrained Video-Editor module and directly training Effect-LoRA causes the generated video to apply particle dispersion effects to all subjects, instead of following the textual instructions to apply them only to the white cake.
Conversely, using only the pretrained Video-Editor without Effect-LoRA fails to produce customized VFX that align with user intent. 
In contrast, the complete method efficiently generates video effects that are highly consistent with the text prompts while faithfully preserving the original video content.
As illustrated in \tabref{tab_ablation}, the performance of the full method across all metrics further confirms these observations.

\section{Conclusion}
\label{sec:conclusion}

In this work, we propose IC-Effect, an instruction-guided few-shot video effect editing framework that learns unique visual styles from a minimal number of paired effect samples. 
By leveraging the intrinsic contextual learning of the DiT architecture, our method performs video VFX editing that strictly follows textual instructions while preserving background consistency. 
The two-stage training strategy ensures both instruction adherence and the learning of specific effect styles. 
Additionally, our spatiotemporal sparse tokenization mechanism preserves full spatiotemporal information from the source video while significantly reducing computational cost. 
We also introduce a new dataset with $15$ types of video effects, providing a valuable benchmark for future research. 
Extensive experiments show that IC-Effect consistently outperforms existing methods in both general video editing and video VFX editing, demonstrating superior effect fidelity and background consistency.

{
    \small
    \bibliographystyle{ieeenat_fullname}
    \bibliography{main}

\begin{thebibliography}{63}
\providecommand{\natexlab}[1]{#1}
\providecommand{\url}[1]{\texttt{#1}}
\expandafter\ifx\csname urlstyle\endcsname\relax
  \providecommand{\doi}[1]{doi: #1}\else
  \providecommand{\doi}{doi: \begingroup \urlstyle{rm}\Url}\fi

\bibitem[gpt()]{gpt4o}
Gpt-4o.
\newblock Accessed May 13, 2024 [Online] \url{https://openai.com/index/hello-gpt-4o/}.

\bibitem[pex()]{pexels}
Pexels.
\newblock \url{https://www.pexels.com/}, 2025.

\bibitem[Batifol et~al.(2025)Batifol, Blattmann, Boesel, Consul, Diagne, Dockhorn, English, English, Esser, Kulal, et~al.]{batifol2025flux}
Stephen Batifol, Andreas Blattmann, Frederic Boesel, Saksham Consul, Cyril Diagne, Tim Dockhorn, Jack English, Zion English, Patrick Esser, Sumith Kulal, et~al.
\newblock Flux. 1 kontext: Flow matching for in-context image generation and editing in latent space.
\newblock \emph{arXiv e-prints}, pages arXiv--2506, 2025.

\bibitem[Belova(2021)]{belova2021google}
Alla Belova.
\newblock Google doodles as multimodal storytelling.
\newblock \emph{Cognition, communication, discourse}, \penalty0 (23):\penalty0 13--29, 2021.

\bibitem[Bian et~al.(2025)Bian, Zhang, Ju, Cao, Xie, Shan, and Xu]{bian2025videopainter}
Yuxuan Bian, Zhaoyang Zhang, Xuan Ju, Mingdeng Cao, Liangbin Xie, Ying Shan, and Qiang Xu.
\newblock Videopainter: Any-length video inpainting and editing with plug-and-play context control.
\newblock In \emph{SIGGRAPH}, pages 1--12, 2025.

\bibitem[Brooks et~al.(2024)Brooks, Peebles, Holmes, DePue, Guo, Jing, Schnurr, Taylor, Luhman, Luhman, Ng, Wang, and Ramesh]{videoworldsimulators2024}
Tim Brooks, Bill Peebles, Connor Holmes, Will DePue, Yufei Guo, Li Jing, David Schnurr, Joe Taylor, Troy Luhman, Eric Luhman, Clarence Ng, Ricky Wang, and Aditya Ramesh.
\newblock Video generation models as world simulators.
\newblock 2024.

\bibitem[Brown et~al.(2020)Brown, Mann, Ryder, Subbiah, Kaplan, Dhariwal, Neelakantan, Shyam, Sastry, and Askell]{NEURIPS2020_1457c0d6}
Tom Brown, Benjamin Mann, Nick Ryder, Melanie Subbiah, Jared~D Kaplan, Prafulla Dhariwal, Arvind Neelakantan, Pranav Shyam, Girish Sastry, and et~al. Askell, Amanda.
\newblock Language models are few-shot learners.
\newblock In \emph{NeurIPS}, pages 1877--1901, 2020.

\bibitem[Chen et~al.(2024)Chen, Zhang, Cun, Xia, Wang, Weng, and Shan]{chen2024videocrafter2}
Haoxin Chen, Yong Zhang, Xiaodong Cun, Menghan Xia, Xintao Wang, Chao Weng, and Ying Shan.
\newblock Videocrafter2: Overcoming data limitations for high-quality video diffusion models.
\newblock In \emph{CVPR}, pages 7310--7320, 2024.

\bibitem[Chen et~al.(2025{\natexlab{a}})Chen, Gu, and Mao]{chen2025univid}
Lan Chen, Yuchao Gu, and Qi Mao.
\newblock Univid: Unifying vision tasks with pre-trained video generation models.
\newblock \emph{arXiv preprint arXiv:2509.21760}, 2025{\natexlab{a}}.

\bibitem[Chen et~al.(2025{\natexlab{b}})Chen, Mao, Gu, and Shou]{chen2025edit}
Lan Chen, Qi Mao, Yuchao Gu, and Mike~Zheng Shou.
\newblock Edit transfer: Learning image editing via vision in-context relations.
\newblock \emph{arXiv preprint arXiv:2503.13327}, 2025{\natexlab{b}}.

\bibitem[Chen et~al.(2025{\natexlab{c}})Chen, Guo, Zhu, Zhang, Huang, Feng, and Kang]{chen2025video}
Sili Chen, Hengkai Guo, Shengnan Zhu, Feihu Zhang, Zilong Huang, Jiashi Feng, and Bingyi Kang.
\newblock Video depth anything: Consistent depth estimation for super-long videos.
\newblock In \emph{CVPR}, pages 22831--22840, 2025{\natexlab{c}}.

\bibitem[Cheng et~al.(2024)Cheng, Xiao, and He]{cheng2024consistent}
Jiaxin Cheng, Tianjun Xiao, and Tong He.
\newblock Consistent video-to-video transfer using synthetic dataset.
\newblock In \emph{ICLR}, 2024.

\bibitem[Esser et~al.(2024)Esser, Kulal, Blattmann, Entezari, M{\"u}ller, Saini, Levi, Lorenz, Sauer, Boesel, et~al.]{esser2024scaling}
Patrick Esser, Sumith Kulal, Andreas Blattmann, Rahim Entezari, Jonas M{\"u}ller, Harry Saini, Yam Levi, Dominik Lorenz, Axel Sauer, Frederic Boesel, et~al.
\newblock Scaling rectified flow transformers for high-resolution image synthesis.
\newblock In \emph{ICML}, 2024.

\bibitem[Gao et~al.(2025)Gao, Ding, Cai, Huang, Wang, and Xue]{gao2025lora}
Chenjian Gao, Lihe Ding, Xin Cai, Zhanpeng Huang, Zibin Wang, and Tianfan Xue.
\newblock Lora-edit: Controllable first-frame-guided video editing via mask-aware lora fine-tuning.
\newblock \emph{arXiv preprint arXiv:2506.10082}, 2025.

\bibitem[Geyer et~al.(2023)Geyer, Bar-Tal, Bagon, and Dekel]{geyer2023tokenflow}
Michal Geyer, Omer Bar-Tal, Shai Bagon, and Tali Dekel.
\newblock Tokenflow: Consistent diffusion features for consistent video editing.
\newblock \emph{arXiv preprint arXiv:2307.10373}, 2023.

\bibitem[Gu et~al.(2024)Gu, Zhou, Wu, Yu, Liu, Zhao, Wu, Zhang, Shou, and Tang]{gu2024videoswap}
Yuchao Gu, Yipin Zhou, Bichen Wu, Licheng Yu, Jia-Wei Liu, Rui Zhao, Jay~Zhangjie Wu, David~Junhao Zhang, Mike~Zheng Shou, and Kevin Tang.
\newblock Videoswap: Customized video subject swapping with interactive semantic point correspondence.
\newblock In \emph{CVPR}, pages 7621--7630, 2024.

\bibitem[Guo et~al.(2023)Guo, Yang, Rao, Liang, Wang, Qiao, Agrawala, Lin, and Dai]{guo2023animatediff}
Yuwei Guo, Ceyuan Yang, Anyi Rao, Zhengyang Liang, Yaohui Wang, Yu Qiao, Maneesh Agrawala, Dahua Lin, and Bo Dai.
\newblock Animatediff: Animate your personalized text-to-image diffusion models without specific tuning.
\newblock \emph{arXiv preprint arXiv:2307.04725}, 2023.

\bibitem[Hertz et~al.(2022)Hertz, Mokady, Tenenbaum, Aberman, Pritch, and Cohen-Or]{hertz2022prompt}
Amir Hertz, Ron Mokady, Jay Tenenbaum, Kfir Aberman, Yael Pritch, and Daniel Cohen-Or.
\newblock Prompt-to-prompt image editing with cross attention control.
\newblock \emph{arXiv preprint arXiv:2208.01626}, 2022.

\bibitem[Hu et~al.(2022)Hu, Wallis, Allen-Zhu, Li, Wang, Wang, Chen, et~al.]{hulora}
Edward~J Hu, Phillip Wallis, Zeyuan Allen-Zhu, Yuanzhi Li, Shean Wang, Lu Wang, Weizhu Chen, et~al.
\newblock Lora: Low-rank adaptation of large language models.
\newblock In \emph{ICLR}, 2022.

\bibitem[Hu and Xu(2023)]{hu2023videocontrolnet}
Zhihao Hu and Dong Xu.
\newblock Videocontrolnet: A motion-guided video-to-video translation framework by using diffusion model with controlnet.
\newblock \emph{arXiv preprint arXiv:2307.14073}, 2023.

\bibitem[Huang et~al.(2024{\natexlab{a}})Huang, Wang, Wu, Shi, Dou, Liang, Feng, Liu, and Zhou]{huang2024context}
Lianghua Huang, Wei Wang, Zhi-Fan Wu, Yupeng Shi, Huanzhang Dou, Chen Liang, Yutong Feng, Yu Liu, and Jingren Zhou.
\newblock In-context lora for diffusion transformers.
\newblock \emph{arXiv preprint arXiv:2410.23775}, 2024{\natexlab{a}}.

\bibitem[Huang et~al.(2025{\natexlab{a}})Huang, Song, Zhang, Guo, Wang, and Liu]{huang2025arteditor}
Shijie Huang, Yiren Song, Yuxuan Zhang, Hailong Guo, Xueyin Wang, and Jiaming Liu.
\newblock Arteditor: Learning customized instructional image editor from few-shot examples.
\newblock In \emph{ICCV}, pages 17651--17662, 2025{\natexlab{a}}.

\bibitem[Huang et~al.(2025{\natexlab{b}})Huang, Song, Zhang, Guo, Wang, Shou, and Liu]{huang2025photodoodle}
Shijie Huang, Yiren Song, Yuxuan Zhang, Hailong Guo, Xueyin Wang, Mike~Zheng Shou, and Jiaming Liu.
\newblock Photodoodle: Learning artistic image editing from few-shot pairwise data.
\newblock \emph{arXiv preprint arXiv:2502.14397}, 2025{\natexlab{b}}.

\bibitem[Huang et~al.(2024{\natexlab{b}})Huang, He, Yu, Zhang, Si, Jiang, Zhang, Wu, Jin, Chanpaisit, et~al.]{huang2024vbench}
Ziqi Huang, Yinan He, Jiashuo Yu, Fan Zhang, Chenyang Si, Yuming Jiang, Yuanhan Zhang, Tianxing Wu, Qingyang Jin, Nattapol Chanpaisit, et~al.
\newblock Vbench: Comprehensive benchmark suite for video generative models.
\newblock In \emph{CVPR}, pages 21807--21818, 2024{\natexlab{b}}.

\bibitem[Jiang et~al.(2025)Jiang, Han, Mao, Zhang, Pan, and Liu]{jiang2025vace}
Zeyinzi Jiang, Zhen Han, Chaojie Mao, Jingfeng Zhang, Yulin Pan, and Yu Liu.
\newblock Vace: All-in-one video creation and editing.
\newblock \emph{arXiv preprint arXiv:2503.07598}, 2025.

\bibitem[Ju et~al.(2025)Ju, Ye, Liu, Wang, Wang, Wan, Zhang, Gai, and Xu]{ju2025fulldit}
Xuan Ju, Weicai Ye, Quande Liu, Qiulin Wang, Xintao Wang, Pengfei Wan, Di Zhang, Kun Gai, and Qiang Xu.
\newblock Fulldit: Video generative foundation models with multimodal control via full attention.
\newblock In \emph{ICCV}, pages 15737--15747, 2025.

\bibitem[Kong et~al.(2024)Kong, Tian, Zhang, Min, Dai, Zhou, Xiong, Li, Wu, Zhang, et~al.]{kong2024hunyuanvideo}
Weijie Kong, Qi Tian, Zijian Zhang, Rox Min, Zuozhuo Dai, Jin Zhou, Jiangfeng Xiong, Xin Li, Bo Wu, Jianwei Zhang, et~al.
\newblock Hunyuanvideo: A systematic framework for large video generative models.
\newblock \emph{arXiv preprint arXiv:2412.03603}, 2024.

\bibitem[Ku et~al.(2024)Ku, Wei, Ren, Yang, and Chen]{ku2024anyvv}
Max Ku, Cong Wei, Weiming Ren, Huan Yang, and Wenhu Chen.
\newblock Anyv2v: A tuning-free framework for any video-to-video editing tasks.
\newblock \emph{TMLR}, 2024.

\bibitem[Li et~al.(2025)Li, Xue, Ren, and Bo]{li2025diffueraser}
Xiaowen Li, Haolan Xue, Peiran Ren, and Liefeng Bo.
\newblock Diffueraser: A diffusion model for video inpainting.
\newblock \emph{arXiv preprint arXiv:2501.10018}, 2025.

\bibitem[Li et~al.(2024)Li, Mao, Chen, Fang, Tian, Xiao, Jin, and Wu]{li2024starvid}
Yuanhang Li, Qi Mao, Lan Chen, Zhen Fang, Lei Tian, Xinyan Xiao, Libiao Jin, and Hua Wu.
\newblock Starvid: Enhancing semantic alignment in video diffusion models via spatial and syntactic guided attention refocusing.
\newblock \emph{arXiv preprint arXiv:2409.15259}, 2024.

\bibitem[Liu et~al.(2024{\natexlab{a}})Liu, Zeng, Ren, Li, Zhang, Yang, Jiang, Li, Yang, Su, et~al.]{liu2024grounding}
Shilong Liu, Zhaoyang Zeng, Tianhe Ren, Feng Li, Hao Zhang, Jie Yang, Qing Jiang, Chunyuan Li, Jianwei Yang, Hang Su, et~al.
\newblock Grounding dino: Marrying dino with grounded pre-training for open-set object detection.
\newblock In \emph{ECCV}, pages 38--55. Springer, 2024{\natexlab{a}}.

\bibitem[Liu et~al.(2024{\natexlab{b}})Liu, Zhang, Li, Lin, and Jia]{liu2024video}
Shaoteng Liu, Yuechen Zhang, Wenbo Li, Zhe Lin, and Jiaya Jia.
\newblock Video-p2p: Video editing with cross-attention control.
\newblock In \emph{CVPR}, pages 8599--8608, 2024{\natexlab{b}}.

\bibitem[Liu et~al.(2025)Liu, Zeng, Xue, Yang, Luo, Liu, and Guo]{liu2025vfx}
Xinyu Liu, Ailing Zeng, Wei Xue, Harry Yang, Wenhan Luo, Qifeng Liu, and Yike Guo.
\newblock Vfx creator: Animated visual effect generation with controllable diffusion transformer.
\newblock \emph{arXiv preprint arXiv:2502.05979}, 2025.

\bibitem[Ma et~al.(2024{\natexlab{a}})Ma, He, Cun, Wang, Chen, Li, and Chen]{ma2024followpose}
Yue Ma, Yingqing He, Xiaodong Cun, Xintao Wang, Siran Chen, Xiu Li, and Qifeng Chen.
\newblock Follow your pose: Pose-guided text-to-video generation using pose-free videos.
\newblock In \emph{AAAI}, pages 4117--4125, 2024{\natexlab{a}}.

\bibitem[Ma et~al.(2024{\natexlab{b}})Ma, Liu, Wang, Pan, He, Yuan, Zeng, Cai, Shum, Liu, et~al.]{ma2024followyouremoji}
Yue Ma, Hongyu Liu, Hongfa Wang, Heng Pan, Yingqing He, Junkun Yuan, Ailing Zeng, Chengfei Cai, Heung-Yeung Shum, Wei Liu, et~al.
\newblock Follow-your-emoji: Fine-controllable and expressive freestyle portrait animation.
\newblock In \emph{SIGGRAPH Asia 2024 Conference Papers}, pages 1--12, 2024{\natexlab{b}}.

\bibitem[Ma et~al.(2025{\natexlab{a}})Ma, He, Wang, Wang, Shen, Qi, Ying, Cai, Li, Shum, et~al.]{ma2025followyourclick}
Yue Ma, Yingqing He, Hongfa Wang, Andong Wang, Leqi Shen, Chenyang Qi, Jixuan Ying, Chengfei Cai, Zhifeng Li, Heung-Yeung Shum, et~al.
\newblock Follow-your-click: Open-domain regional image animation via motion prompts.
\newblock In \emph{AAAI}, pages 6018--6026, 2025{\natexlab{a}}.

\bibitem[Ma et~al.(2025{\natexlab{b}})Ma, Liu, Zhu, Yang, Feng, Zhang, Li, Han, Qi, and Chen]{ma2025followyourmotion}
Yue Ma, Yulong Liu, Qiyuan Zhu, Ayden Yang, Kunyu Feng, Xinhua Zhang, Zhifeng Li, Sirui Han, Chenyang Qi, and Qifeng Chen.
\newblock Follow-your-motion: Video motion transfer via efficient spatial-temporal decoupled finetuning.
\newblock \emph{arXiv preprint arXiv:2506.05207}, 2025{\natexlab{b}}.

\bibitem[Mao et~al.(2025)Mao, Hao, Chen, Liu, Feng, Zhu, Wu, Chen, Wu, and Chu]{mao2025omni}
Fangyuan Mao, Aiming Hao, Jintao Chen, Dongxia Liu, Xiaokun Feng, Jiashu Zhu, Meiqi Wu, Chubin Chen, Jiahong Wu, and Xiangxiang Chu.
\newblock Omni-effects: Unified and spatially-controllable visual effects generation.
\newblock \emph{arXiv preprint arXiv:2508.07981}, 2025.

\bibitem[Mao et~al.(2024)Mao, Chen, Gu, Fang, and Shou]{mao2024mag}
Qi Mao, Lan Chen, Yuchao Gu, Zhen Fang, and Mike~Zheng Shou.
\newblock Mag-edit: Localized image editing in complex scenarios via mask-based attention-adjusted guidance.
\newblock In \emph{ACM MM}, pages 6842--6850, 2024.

\bibitem[Peebles and Xie(2023)]{peebles2023scalable}
William Peebles and Saining Xie.
\newblock Scalable diffusion models with transformers.
\newblock In \emph{CVPR}, pages 4195--4205, 2023.

\bibitem[Pont-Tuset et~al.(2017)Pont-Tuset, Perazzi, Caelles, Arbel{\'a}ez, Sorkine-Hornung, and Van~Gool]{pont20172017}
Jordi Pont-Tuset, Federico Perazzi, Sergi Caelles, Pablo Arbel{\'a}ez, Alex Sorkine-Hornung, and Luc Van~Gool.
\newblock The 2017 davis challenge on video object segmentation.
\newblock \emph{arXiv preprint arXiv:1704.00675}, 2017.

\bibitem[Radford et~al.(2021)Radford, Kim, Hallacy, Ramesh, Goh, Agarwal, Sastry, Askell, Mishkin, Clark, et~al.]{radford2021learning}
Alec Radford, Jong~Wook Kim, Chris Hallacy, Aditya Ramesh, Gabriel Goh, Sandhini Agarwal, Girish Sastry, Amanda Askell, Pamela Mishkin, Jack Clark, et~al.
\newblock Learning transferable visual models from natural language supervision.
\newblock In \emph{ICML}, pages 8748--8763. PMLR, 2021.

\bibitem[Ravi et~al.(2024)Ravi, Gabeur, Hu, Hu, Ryali, Ma, Khedr, R{\"a}dle, Rolland, Gustafson, et~al.]{ravi20242}
Nikhila Ravi, Valentin Gabeur, Yuan-Ting Hu, Ronghang Hu, Chaitanya Ryali, Tengyu Ma, H Khedr, R R{\"a}dle, C Rolland, L Gustafson, et~al.
\newblock Sam 2: Segment anything in images and videos.
\newblock \emph{arXiv preprint arXiv:2408.00714}, 2024.

\bibitem[Rombach et~al.(2022)Rombach, Blattmann, Lorenz, Esser, and Ommer]{rombach2022high}
Robin Rombach, Andreas Blattmann, Dominik Lorenz, Patrick Esser, and Bj{\"o}rn Ommer.
\newblock High-resolution image synthesis with latent diffusion models.
\newblock In \emph{CVPR}, pages 10684--10695, 2022.

\bibitem[Song et~al.(2024)Song, Huang, Yao, Ye, Ci, Liu, Zhang, and Shou]{song2024processpainter}
Yiren Song, Shijie Huang, Chen Yao, Xiaojun Ye, Hai Ci, Jiaming Liu, Yuxuan Zhang, and Mike~Zheng Shou.
\newblock Processpainter: Learn painting process from sequence data.
\newblock \emph{arXiv preprint arXiv:2406.06062}, 2024.

\bibitem[Song et~al.(2025{\natexlab{a}})Song, Chen, and Shou]{song2025layertracer}
Yiren Song, Danze Chen, and Mike~Zheng Shou.
\newblock Layertracer: Cognitive-aligned layered svg synthesis via diffusion transformer.
\newblock \emph{arXiv preprint arXiv:2502.01105}, 2025{\natexlab{a}}.

\bibitem[Song et~al.(2025{\natexlab{b}})Song, Liu, and Shou]{song2025makeanything}
Yiren Song, Cheng Liu, and Mike~Zheng Shou.
\newblock Makeanything: Harnessing diffusion transformers for multi-domain procedural sequence generation.
\newblock \emph{arXiv preprint arXiv:2502.01572}, 2025{\natexlab{b}}.

\bibitem[Song et~al.(2025{\natexlab{c}})Song, Liu, and Shou]{song2025omniconsistency}
Yiren Song, Cheng Liu, and Mike~Zheng Shou.
\newblock Omniconsistency: Learning style-agnostic consistency from paired stylization data.
\newblock \emph{arXiv preprint arXiv:2505.18445}, 2025{\natexlab{c}}.

\bibitem[Tan et~al.(2025)Tan, Xue, Yang, Liu, and Wang]{tan2025ominicontrol2}
Zhenxiong Tan, Qiaochu Xue, Xingyi Yang, Songhua Liu, and Xinchao Wang.
\newblock Ominicontrol2: Efficient conditioning for diffusion transformers.
\newblock \emph{arXiv preprint arXiv:2503.08280}, 2025.

\bibitem[Team(2025)]{decart2025lucyedit}
DecartAI Team.
\newblock Lucy edit: Open-weight text-guided video editing.
\newblock 2025.

\bibitem[Teed and Deng(2020)]{teed2020raft}
Zachary Teed and Jia Deng.
\newblock Raft: Recurrent all-pairs field transforms for optical flow.
\newblock In \emph{ECCV}, pages 402--419. Springer, 2020.

\bibitem[Wan et~al.(2025)Wan, Wang, Ai, Wen, Mao, Xie, Chen, Yu, Zhao, Yang, et~al.]{wan2025wan}
Team Wan, Ang Wang, Baole Ai, Bin Wen, Chaojie Mao, Chen-Wei Xie, Di Chen, Feiwu Yu, Haiming Zhao, Jianxiao Yang, et~al.
\newblock Wan: Open and advanced large-scale video generative models.
\newblock \emph{arXiv preprint arXiv:2503.20314}, 2025.

\bibitem[Wang et~al.(2023)Wang, Yuan, Chen, Zhang, Wang, and Zhang]{wang2023modelscope}
Jiuniu Wang, Hangjie Yuan, Dayou Chen, Yingya Zhang, Xiang Wang, and Shiwei Zhang.
\newblock Modelscope text-to-video technical report.
\newblock \emph{arXiv preprint arXiv:2308.06571}, 2023.

\bibitem[Wang et~al.(2024)Wang, He, Li, Li, Yu, Ma, Li, Chen, Chen, Wang, Luo, Liu, Wang, Wang, and Qiao]{wang2024internvid}
Yi Wang, Yinan He, Yizhuo Li, Kunchang Li, Jiashuo Yu, Xin Ma, Xinhao Li, Guo Chen, Xinyuan Chen, Yaohui Wang, Ping Luo, Ziwei Liu, Yali Wang, Limin Wang, and Yu Qiao.
\newblock Internvid: A large-scale video-text dataset for multimodal understanding and generation.
\newblock In \emph{ICLR}, 2024.

\bibitem[Wu et~al.(2023)Wu, Ge, Wang, Lei, Gu, Shi, Hsu, Shan, Qie, and Shou]{wu2023tune}
Jay~Zhangjie Wu, Yixiao Ge, Xintao Wang, Stan~Weixian Lei, Yuchao Gu, Yufei Shi, Wynne Hsu, Ying Shan, Xiaohu Qie, and Mike~Zheng Shou.
\newblock Tune-a-video: One-shot tuning of image diffusion models for text-to-video generation.
\newblock In \emph{ICCV}, pages 7623--7633, 2023.

\bibitem[Wu et~al.(2025)Wu, Chen, Li, Wang, Xie, and Zhang]{wu2025insvie}
Yuhui Wu, Liyi Chen, Ruibin Li, Shihao Wang, Chenxi Xie, and Lei Zhang.
\newblock Insvie-1m: Effective instruction-based video editing with elaborate dataset construction.
\newblock In \emph{ICCV}, pages 16692--16701, 2025.

\bibitem[Yang et~al.(2025)Yang, Teng, Zheng, Ding, Huang, Xu, Yang, Hong, Zhang, Feng, Yin, Yuxuan.Zhang, Wang, Cheng, Xu, Gu, Dong, and Tang]{yang2025cogvideox}
Zhuoyi Yang, Jiayan Teng, Wendi Zheng, Ming Ding, Shiyu Huang, Jiazheng Xu, Yuanming Yang, Wenyi Hong, Xiaohan Zhang, Guanyu Feng, Da Yin, Yuxuan.Zhang, Weihan Wang, Yean Cheng, Bin Xu, Xiaotao Gu, Yuxiao Dong, and Jie Tang.
\newblock Cogvideox: Text-to-video diffusion models with an expert transformer.
\newblock In \emph{ICLR}, 2025.

\bibitem[Ye et~al.(2025)Ye, He, Liu, Wang, Wang, Wan, Zhang, Gai, Chen, and Luo]{ye2025unic}
Zixuan Ye, Xuanhua He, Quande Liu, Qiulin Wang, Xintao Wang, Pengfei Wan, Di Zhang, Kun Gai, Qifeng Chen, and Wenhan Luo.
\newblock Unic: Unified in-context video editing.
\newblock \emph{arXiv preprint arXiv:2506.04216}, 2025.

\bibitem[Yu et~al.(2023)Yu, Blackburn-Matzen, Nguyen, Wang, Habib~Kazi, and Bousseau]{yu2023videodoodles}
Emilie Yu, Kevin Blackburn-Matzen, Cuong Nguyen, Oliver Wang, Rubaiat Habib~Kazi, and Adrien Bousseau.
\newblock Videodoodles: Hand-drawn animations on videos with scene-aware canvases.
\newblock \emph{ACM Transactions on Graphics (TOG)}, 42\penalty0 (4):\penalty0 1--12, 2023.

\bibitem[Zhang et~al.(2024{\natexlab{a}})Zhang, Song, Liu, Wang, Yu, Tang, Li, Tang, Hu, Pan, et~al.]{zhang2024ssr}
Yuxuan Zhang, Yiren Song, Jiaming Liu, Rui Wang, Jinpeng Yu, Hao Tang, Huaxia Li, Xu Tang, Yao Hu, Han Pan, et~al.
\newblock Ssr-encoder: Encoding selective subject representation for subject-driven generation.
\newblock In \emph{CVPR}, pages 8069--8078, 2024{\natexlab{a}}.

\bibitem[Zhang et~al.(2024{\natexlab{b}})Zhang, Wei, Zhang, Song, Liu, Li, Tang, Hu, and Zhao]{zhang2024stable}
Yuxuan Zhang, Lifu Wei, Qing Zhang, Yiren Song, Jiaming Liu, Huaxia Li, Xu Tang, Yao Hu, and Haibo Zhao.
\newblock Stable-makeup: When real-world makeup transfer meets diffusion model.
\newblock \emph{arXiv preprint arXiv:2403.07764}, 2024{\natexlab{b}}.

\bibitem[Zhang et~al.(2025{\natexlab{a}})Zhang, Yuan, Song, Wang, and Liu]{zhang2025easycontrol}
Yuxuan Zhang, Yirui Yuan, Yiren Song, Haofan Wang, and Jiaming Liu.
\newblock Easycontrol: Adding efficient and flexible control for diffusion transformer.
\newblock In \emph{ICCV}, pages 19513--19524, 2025{\natexlab{a}}.

\bibitem[Zhang et~al.(2025{\natexlab{b}})Zhang, Zhang, Song, Zhang, Tang, and Liu]{zhang2025stable}
Yuxuan Zhang, Qing Zhang, Yiren Song, Jichao Zhang, Hao Tang, and Jiaming Liu.
\newblock Stable-hair: Real-world hair transfer via diffusion model.
\newblock In \emph{AAAI}, pages 10348--10356, 2025{\natexlab{b}}.

\end{thebibliography}
}

\appendix
\clearpage
\setcounter{page}{1}
\maketitlesupplementary

In this supplementary material, we provide additional implementation details, extended ablation studies, more qualitative results, and a discussion of limitations, as summarized below:

\begin{compactitem}
\item In \secref{imple_details}, we present additional implementation details of IC-Effect, the baselines, the quantitative metrics, and the user study setup.

\item In \secref{dataset}, we describe the datasets used for Video-Editor pretraining as well as the paired VFX editing dataset.

\item In \secref{multi_prompt} and \secref{multi_effect}, we demonstrate that IC-Effect supports flexible instruction control and multi-effect editing.

\item In \secref{ablation}, we conduct ablation studies on causal attention and positional correction.

\item In \secref{limitation}, we discuss the limitations of the proposed method and outline future directions.

\item In \secref{add_result}, we provide additional video VFX editing results of IC-Effect and present extended qualitative comparisons with baseline methods on both general video editing and video VFX editing tasks.

\end{compactitem}

\section{Implementation Details}
\label{sec:imple_details}

\subsection{Training Details}

During the pre-training stage of VideoEditor, we initialize the DiT architecture using the Wan 2.2-A14B-T2V\footnote{\url{https://huggingface.co/Wan-AI/Wan2.2-I2V-A14B-Diffusers}} weights from the Wan model. 
All training videos are resized to a spatial resolution of $224 \times 416$ with $81$ frames.
We train a rank-$96$ LoRA for $50$k iterations using the AdamW optimizer with a batch size of $2$, a learning rate of $1 \times 10^{-4}$, and a weight decay of $1 \times 10^{-2}$, on four A800 GPUs.
During the Effect-LoRA training stage, we initialize the DiT architecture with the merged weights from VideoEditor. 
The video resolution and frame count remain $224 \times 416$ and $81$ frames, respectively.
We train a rank-$32$ LoRA for $1000$ iterations using AdamW with a batch size of $2$, a learning rate of $1 \times 10^{-4}$, and a weight decay of $1 \times 10^{-2}$, on two A800 GPUs.

\subsection{Inference Details}

During inference, we generate videos at a spatial resolution of $480 \times 832$. 
For source videos with $\ge 81$ frames, we select the first $81$ frames; for videos with $< 81$ frames, we select the first $4n+1$ frames as the source.
We set the classifier-free guidance scale to $5.0$ and perform $50$ denoising steps.

\subsection{Implementation Details of Baselines}

For common video editing comparisons, we use the official codes and released weights of InsV2V~\cite{cheng2024consistent}\footnote{\url{https://github.com/amazon-science/instruct-video-to-video}}, InsViE~\cite{wu2025insvie}\footnote{\url{https://github.com/langmanbusi/InsViE}}, VACE~\cite{jiang2025vace}\footnote{\url{https://github.com/ali-vilab/VACE}}, and Lucy Edit~\cite{decart2025lucyedit}\footnote{\url{https://github.com/DecartAI/lucy-edit-comfyui}}. 
For video VFX editing comparisons, we train the competing methods on the same paired VFX dataset using their officially provided training code or our faithful re-implementations.

\subsection{Evaluation Details}

\Paragraph{Automatic Evaluation.}
We use the CLIP Vit-L/14~\cite{radford2021learning}\footnote{\url{https://huggingface.co/openai/clip-vit-large-patch14}} model to calculate CLIP Image Similarity and CLIP Text Alignment, assessing the quality of the edited videos in terms of both temporal consistency and alignment with the textual prompts. 
We further evaluate the semantic consistency between the edited videos and their corresponding textual prompts by computing video-level semantic similarity using ViClip~\cite{wang2024internvid}\footnote{\url{https://huggingface.co/OpenGVLab/ViCLIP}}.
In addition, we adopt multiple sub-metrics from VBench, including motion smoothness, dynamic degree, and aesthetic quality, to comprehensively assess the visual fidelity and naturalness of the edited videos. 
Following the official VBench settings, we compute motion smoothness using a frame interpolation model\footnote{\url{https://huggingface.co/lalala125/AMT/tree/main}}, evaluate dynamic degree using RAFT~\cite{teed2020raft}\footnote{\url{https://dl.dropboxusercontent.com/s/4j4z58wuv8o0mfz/models.zip}}, and assess aesthetic quality using the LAION aesthetic predictor\footnote{\url{https://github.com/LAION-AI/aesthetic-predictor}}.

\begin{figure*}[!t]
    \centering
    \includegraphics[width=\linewidth]{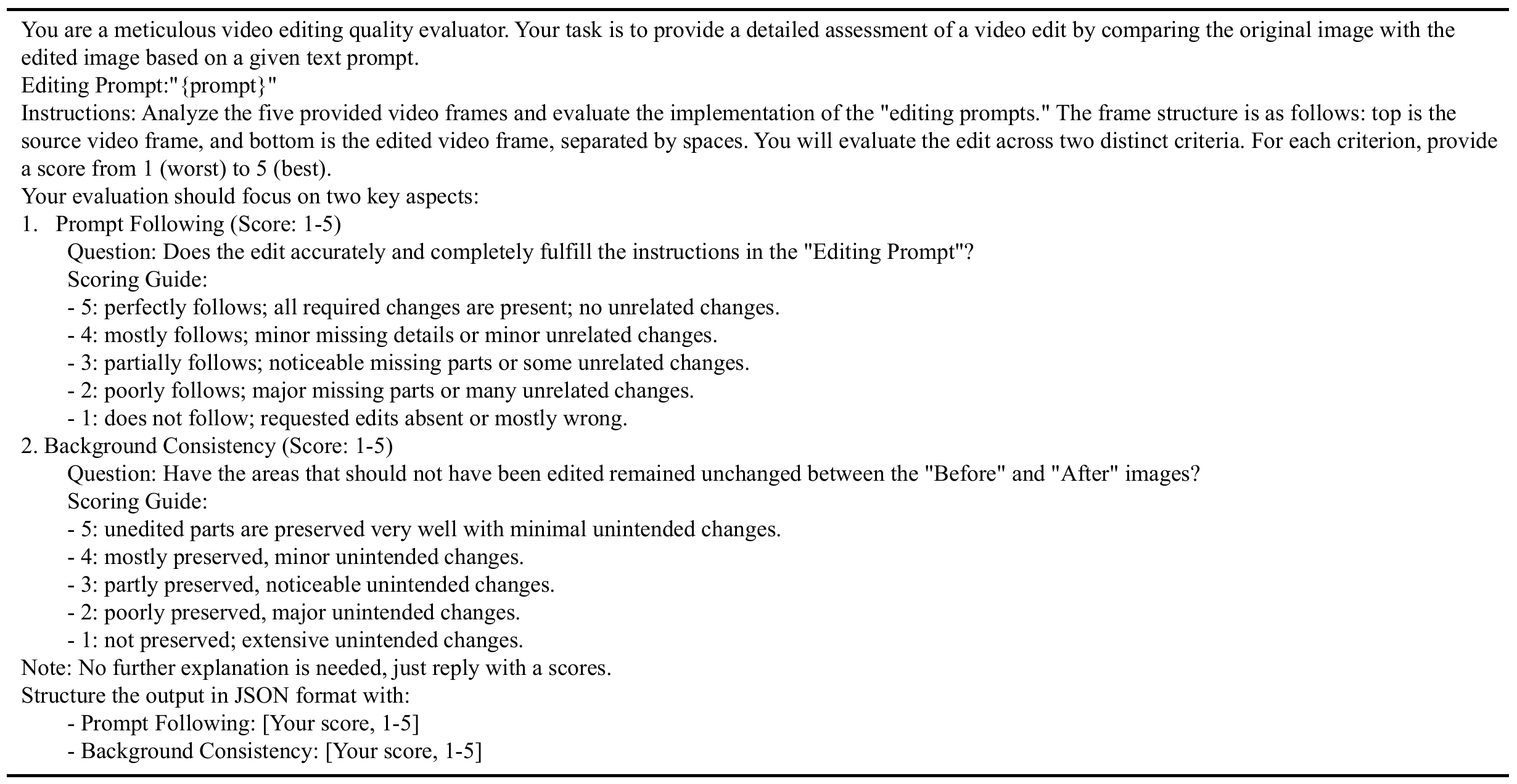}    
    \caption{
    \textbf{Prompt Template for Common Video Editing Evaluation.} 
    }
    \label{fig:LLM_score_common}
\end{figure*}

\begin{figure*}[!t]
    \centering
    \includegraphics[width=\linewidth]{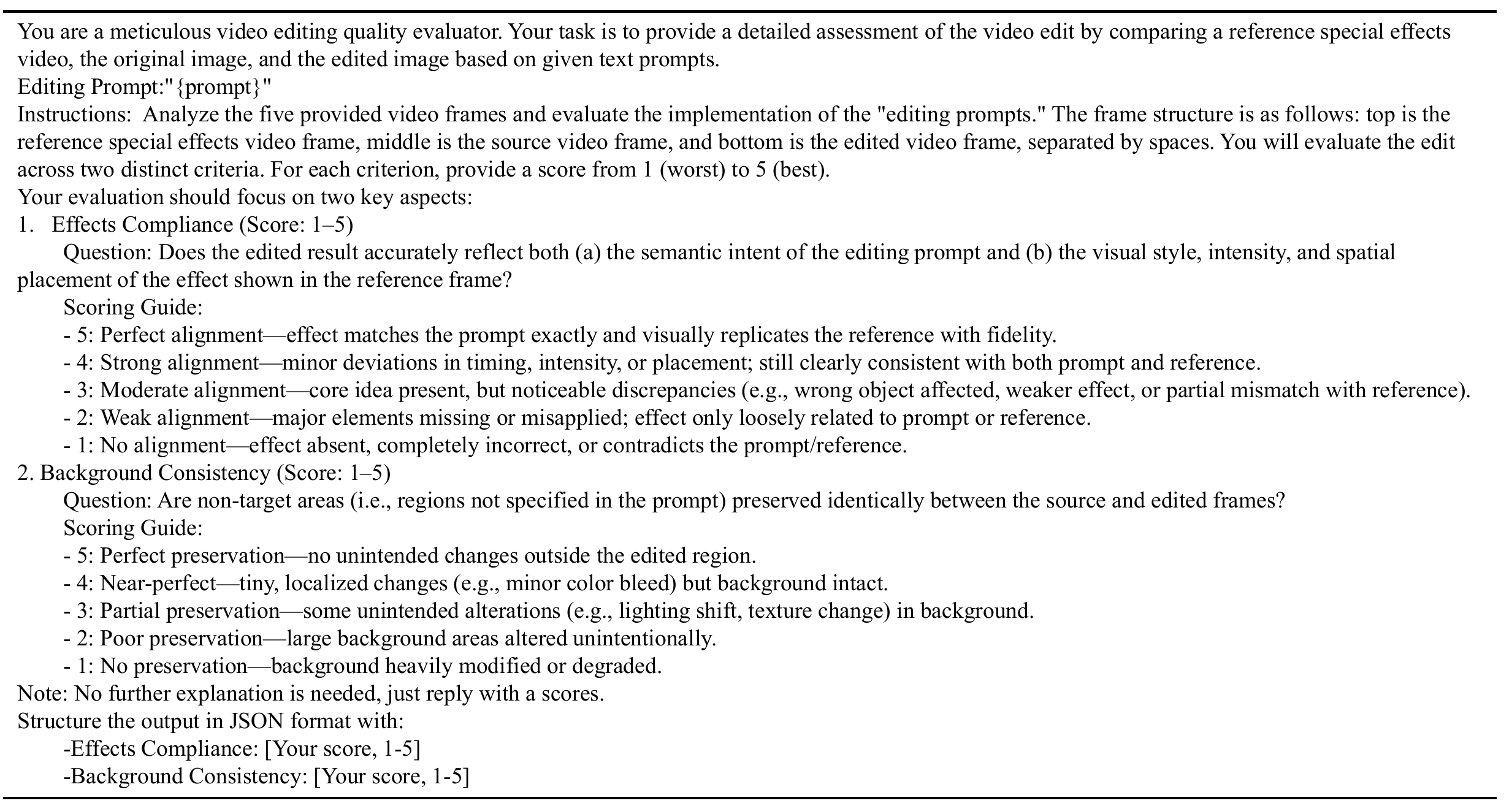}    
    \caption{
    \textbf{Prompt Template for Video VFX Editing Evaluation.} 
    }
    \label{fig:LLM_score_vfx}
\end{figure*}

\Paragraph{GPT-4o Evaluation.}
 To evaluate the structural preservation and editing effectiveness of the generated results, we further employ the state-of-the-art vision–language model (VLM) GPT-4o~\cite{gpt4o} as an automatic evaluator. 
 For general video editing, we uniformly sample five frames from both the source video and the edited video. 
 For each sample, the VLM receives the source frames, the edited frames, and the textual instruction, and it is required to score the editing result on two dimensions—structural preservation and editing effectiveness—using a $1-5$ rating scale.
For video VFX editing, we additionally provide the reference VFX video so that the VLM can assess editing effectiveness with respect to the visual effect demonstrated in the reference. 
\figref{LLM_score_common} and \figref{LLM_score_vfx} present the prompt templates used to guide the VLM during automatic evaluation.

\Paragraph{User Study.}
For common video editing comparisons, each questionnaire presents participants with the source video, the edited videos produced by the competing methods, our edited video, and the corresponding editing instruction. 
The edited videos from the competing methods and ours are randomly ordered to prevent participants from inferring their origins. 
We ask participants to answer questions along the following three dimensions:

\begin{compactitem}
\item \textbf{Instruction Following}: Which edited video better follows the textual instruction?
\item \textbf{Structual Fidelity}: Which edited video preserves the non-edited regions more faithfully and aligns better with the source video?
\item \textbf{Overall Preference}: Which edited result do you prefer subjectively?
\end{compactitem}

For video VFX editing comparisons, we additionally present the corresponding reference VFX video and ask participants to answer questions along the following three dimensions:

\begin{compactitem}
\item \textbf{Effect Consistency}: Which edited video better follows the textual instruction and more closely matches the visual effect in the reference VFX video?
\item \textbf{Structual Fidelity}: Which edited video preserves the non-edited regions more faithfully and aligns better with the source video?
\item \textbf{Overall Preference}: Which edited result do you prefer subjectively?
\end{compactitem}

\section{Datasets}
\label{sec:dataset}

\subsection{Datasets of Pretrained Video-Editor}

To pre-train VideoEditor and equip it with strong instruction-following capability, we construct a high-quality paired video editing dataset. 
Considering the requirements of video VFX editing, the constructed dataset covers the following editing tasks: addition, removal, replacement, attribute modification, and style transfer. 
The source videos primarily come from high-resolution videos on Pexels~\cite{pexels}.

For addition and removal data, we use GPT-4o~\cite{gpt4o} to analyze each video and identify the target object category. 
We then obtain the corresponding object masks using Grounding DINO~\cite{liu2024grounding} and SAM~\cite{ravi20242}. 
With the masks, we remove the target object using the object erasing model DiffuEraser~\cite{li2025diffueraser}. 
The original video and the object-removed video form a pair for both addition and removal tasks, and we use GPT-4o~\cite{gpt4o} to generate the corresponding “remove” and “add” editing instructions. 
To ensure both video quality and instruction correctness, we further refine the data using a VLM and manual verification.

For replacement and attribute modification data, we employ GPT-4o~\cite{gpt4o} to generate editing instructions for each source video, and then synthesize the edited video using VACE~\cite{jiang2025vace}. We filter the generated videos using a VLM to remove blurry or invalid results.

For style transfer data, GPT-4o~\cite{gpt4o} first generates the textual editing descriptions. We then edit the first frame using FLUX-kontext~\cite{batifol2025flux}, estimate depth maps of the source video using a video depth estimator~\cite{chen2025video}, and feed the edited first frame, depth maps, and textual prompt into VACE~\cite{jiang2025vace} to obtain the final stylized video.

In total, we collect approximately 50,000 high-quality paired video editing samples. All videos are resized to a spatial resolution of $480 \times 832$.

\subsection{Video VFX Datasets}

Since existing techniques cannot automatically produce high-quality paired video VFX editing data, we construct the first paired dataset consisting of source videos and their corresponding VFX videos. 
The dataset includes 15 distinct visual effects, covering: animated character insertion (with two different characters), anime duplication, rotating rings, graffiti strokes, lightning outlines, mystical aura, firework explosion, flame burning, architectural growth and bounce, line traversal, line shaping, particle dispersion, particle aggregation, and light-particle traversal. 
For each pair, we provide a \{source video, target VFX video, editing instruction\} triplet. 
All videos are standardized to a resolution of $480 \times 832$ and a duration of $5$ seconds at $16$ FPS.

\begin{figure}[!t]
    \centering
    \includegraphics[width=\linewidth]{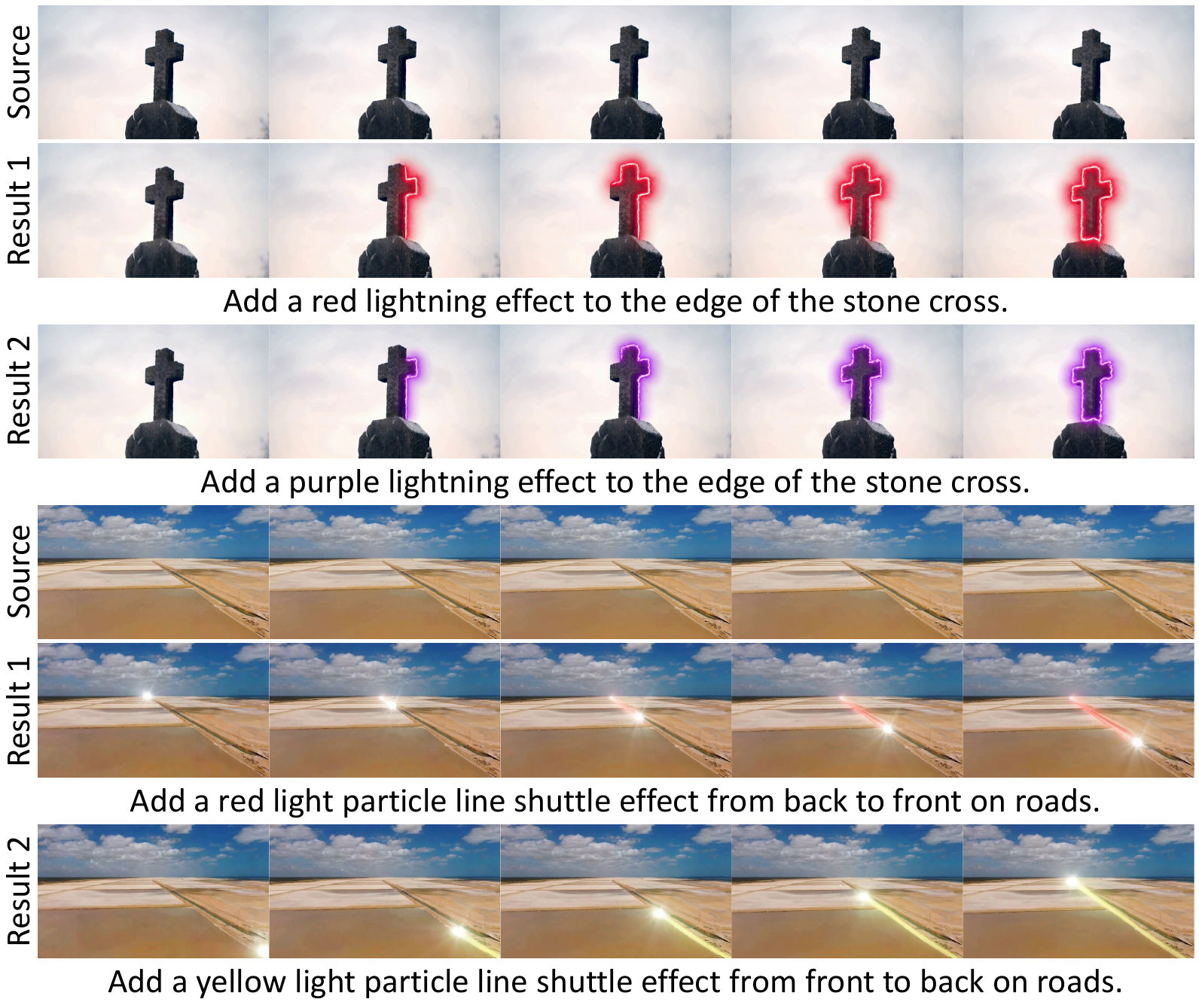}    
    \vspace{-7 mm}
    \caption{
    \textbf{Flexible Instruction-Based Control of Video VFX Editing Results.} 
    }
    \vspace{-5 mm}
    \label{fig:multi_prompt}
\end{figure}

\begin{figure}[!t]
    \centering
    \includegraphics[width=\linewidth]{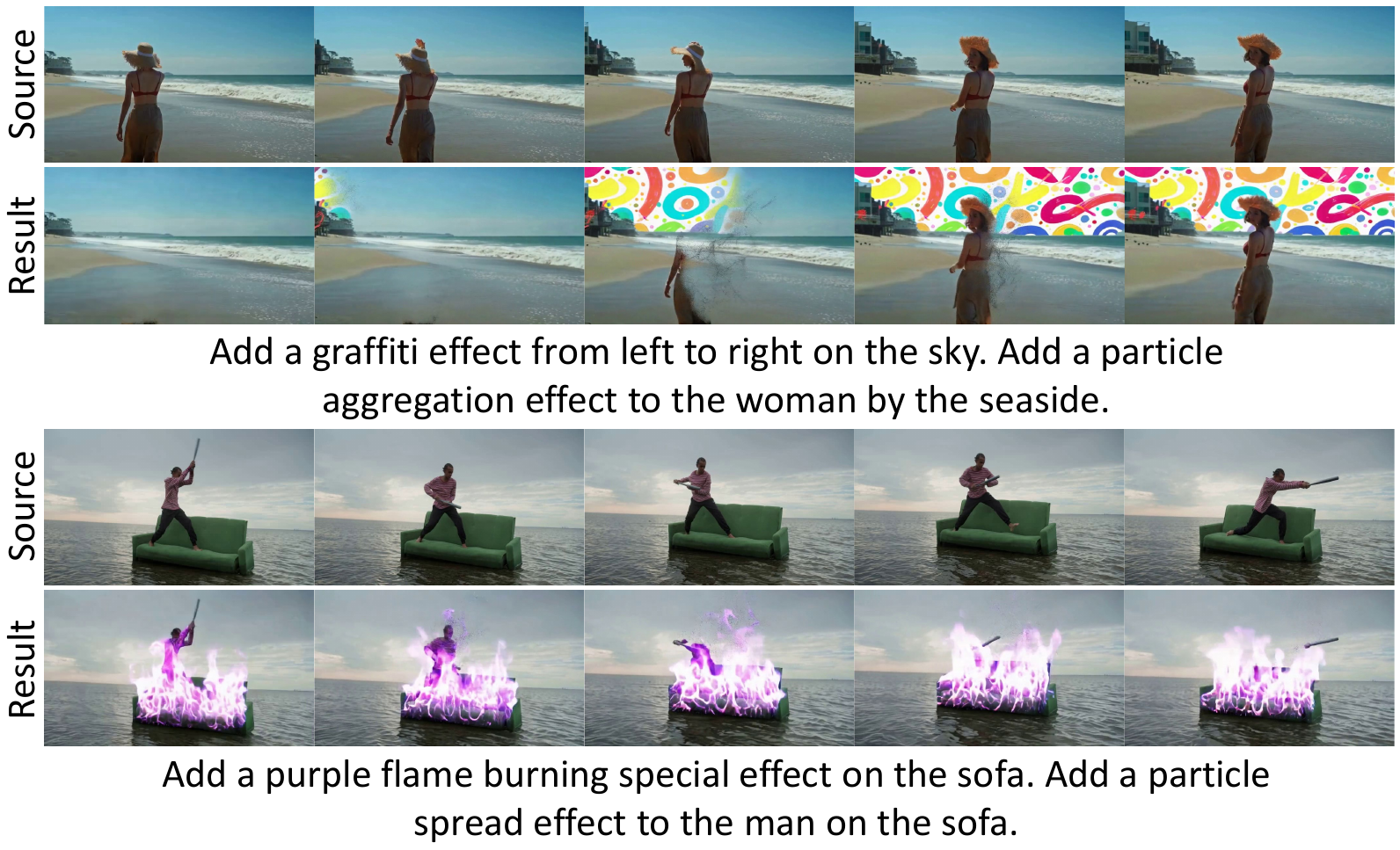}    
    \vspace{-7 mm}
    \caption{
    \textbf{Video Multi VFX Editing of IC-Effect.} 
    }
    \vspace{-5 mm}
    \label{fig:multi_effect}
\end{figure}

\begin{table*}[!t]
\centering
\caption{\textbf{Quantitative Ablation of Positional Encoding Correction and Causal Attention.}
The best values are highlighted in \colorbox{pearDark!20}{blue}.
}
\vspace{-3 mm}
\resizebox{\linewidth}{!}{
\begin{tabular}{ccccccccc}
\hline
\multicolumn{1}{c}{} & \multicolumn{1}{c}{\textbf{Video Quality}}  & \multicolumn{2}{c}{\textbf{Semantic Alignment}} & \multicolumn{3}{c}{\textbf{Overall Quality}} & \multicolumn{2}{c}{\textbf{GPT Score}}  \\ \cmidrule(r){2-2} \cmidrule(r){3-4} \cmidrule(r){5-7} \cmidrule(r){8-9} 
\multicolumn{1}{c}{\multirow{-2}{*}{\textbf{Method/Metrics}}} & \textbf{CLIP-I} \textbf{($\uparrow$)} & \textbf{CLIP-T} \textbf{($\uparrow$)} & \textbf{ViCLIP-T} \textbf{($\uparrow$)} & \textbf{Smoothness} \textbf{($\uparrow$)} & \textbf{Dynamic Degree} \textbf{($\uparrow$)} & \textbf{Aesthetic Quality} \textbf{($\uparrow$)} & \textbf{Structural preservation} \textbf{($\uparrow$)} & \textbf{Effect accuracy} \textbf{($\uparrow$)}\\ 
\midrule
 w/o PEC  &  0.9735 & 26.8748 &  25.7468 &  0.3715 &  0.9903  &  0.5324 & 4.4615 & 4.3908    \\
 w/o C-Attn &   0.9783 & 27.0595 & 26.4503 &  0.3750 &  0.9887  &  0.5444 & 4.5583 &  4.4866    \\
  \textbf{Ours} &   \colorbox{pearDark!20}{0.9786} &   \colorbox{pearDark!20}{27.2321} &  \colorbox{pearDark!20}{26.6312}  & \colorbox{pearDark!20}{0.9911} &  \colorbox{pearDark!20}{0.3771} &   \colorbox{pearDark!20}{0.5823} &   \colorbox{pearDark!20}{4.7947}  & \ \colorbox{pearDark!20}{4.5614}   \\
\bottomrule
\end{tabular}
}
\vspace{-5 mm}
\label{tab:tab_ablation_2}
\end{table*}

\section{Instruction Control for Video VFX Editing}
\label{sec:multi_prompt}

Our IC-Effect framework supports flexible control over the editing results through instruction manipulation. 
By simply adjusting the textual prompt, the model flexibly controls the attributes and directions of the generated visual effects. 
As shown in \figref{multi_prompt}, minor modifications to the prompt already produce clear and controllable variations in the edited effects.

\section{Video Multi VFX Editing}
\label{sec:multi_effect}

To verify that IC-Effect supports multi-effect editing, we perform mixed training on all effect categories based on Video-Editor. 
As shown in \figref{multi_effect}, IC-Effect accurately follows the textual instructions to inject multiple effects into the video without cross-effect leakage. 
It applies each effect precisely to the instruction-specified target while consistently preserving the non-edited regions. 
This capability mainly stems from the carefully designed architecture and the construction of high-quality training data, which together endow IC-Effect with strong instruction-following ability and enable precise source-video editing.

\section{Additional Ablation Studies}
\label{sec:ablation}

\begin{figure}[!t]
    \centering
    \includegraphics[width=\linewidth]{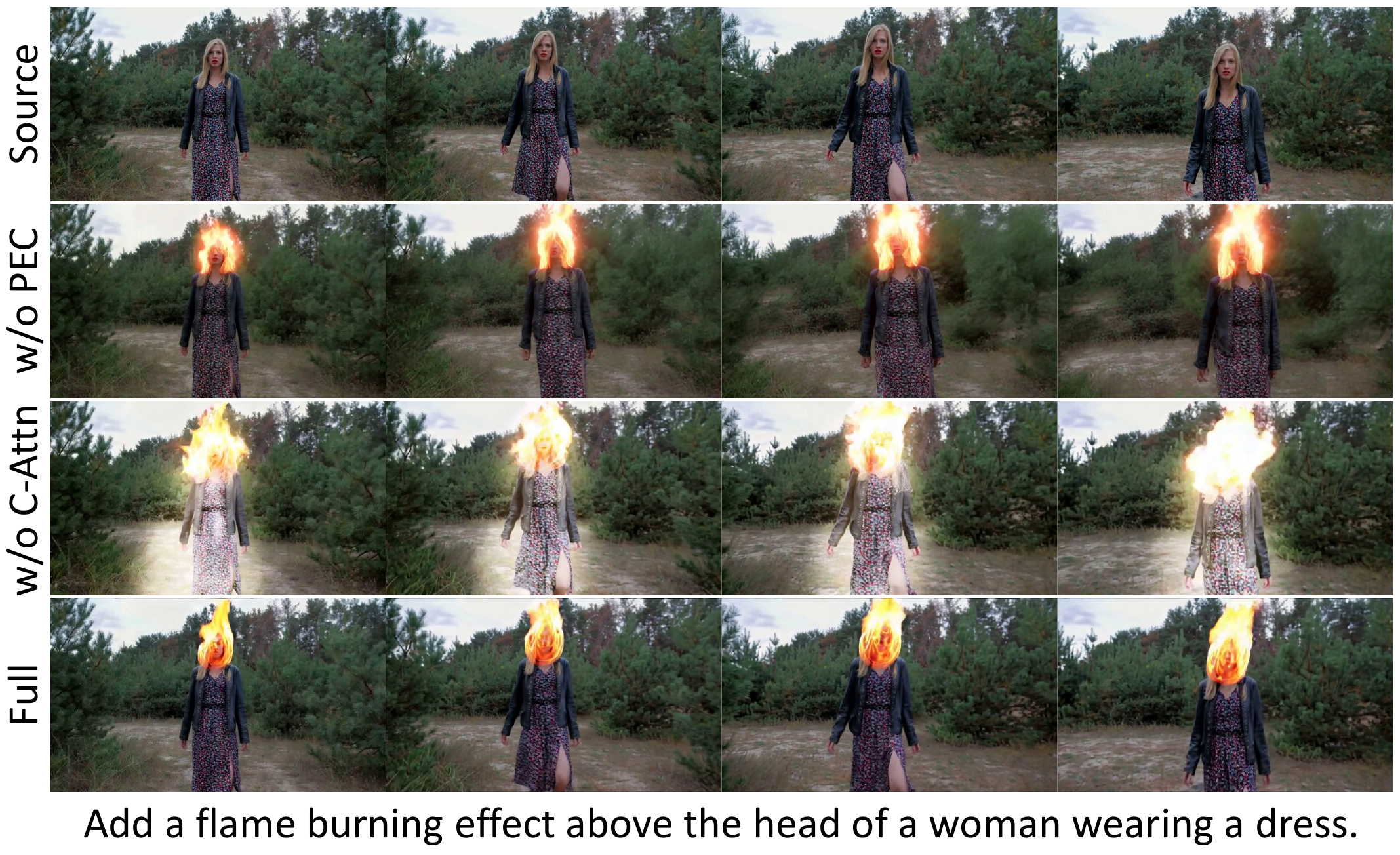}    
    \vspace{-5 mm}
    \caption{
    \textbf{Ablation Study on Positional Encoding Correction and Causal Attention.} 
    }
    \vspace{-5 mm}
    \label{fig:ablation_add}
\end{figure}

In this section, we conduct ablation studies on the positional encoding correction (PEC) and the causal attention mechanism (C-Attn). 
As shown in \figref{ablation_add}, removing positional encoding correction introduces artifacts and blurring, while altering information from the source video. 
Without the causal attention mechanism, the originally clear spatiotemporally sparse tokens are contaminated by latent noise, resulting in severe artifacts in the edited output. 
Furthermore, the quantitative results in \tabref{tab_ablation_2} corroborate these observations. 
In contrast, the complete model demonstrates higher editing accuracy and stronger visual consistency.

\section{Limitation and Future Work}
\label{sec:limitation}

A limitation of IC-Effect is its reliance on high-quality paired video VFX editing data. However, producing such high-quality paired data is extremely challenging. 
In the future, we plan to train a feature extractor to directly extract target effects from reference VFX videos and use these extracted features to edit the source video.

\section{Additional Results}
\label{sec:add_result}

\subsection{Additional Results of Video VFX Editing}

\begin{figure*}[!t]
    \centering
    \includegraphics[width=\linewidth]{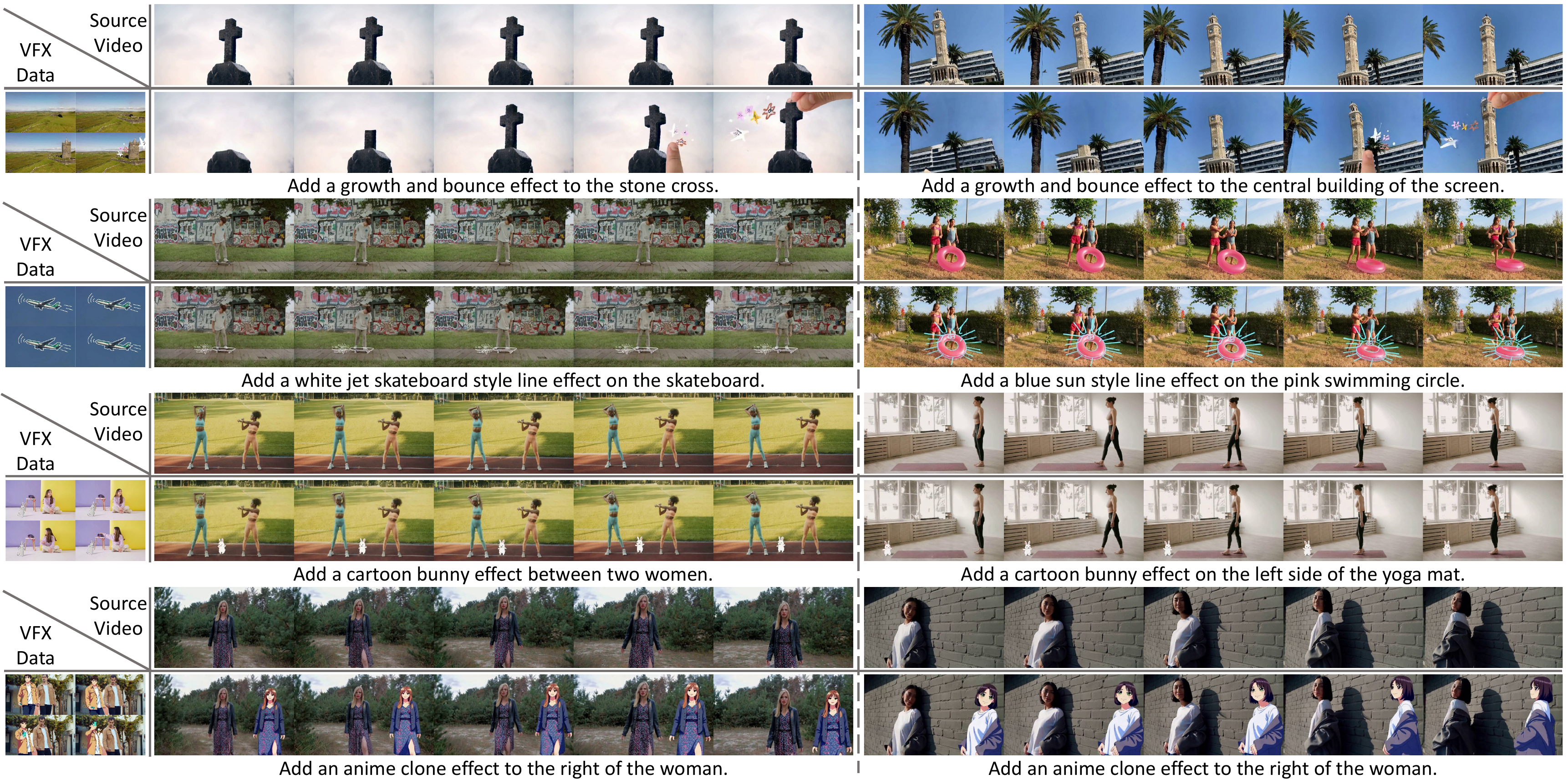}    
    \caption{
    \textbf{Additional Video VFX Editing Results of IC-Effect.} 
    }
    \label{fig:result_0}
\end{figure*}

\begin{figure*}[!t]
    \centering
    \includegraphics[width=\linewidth]{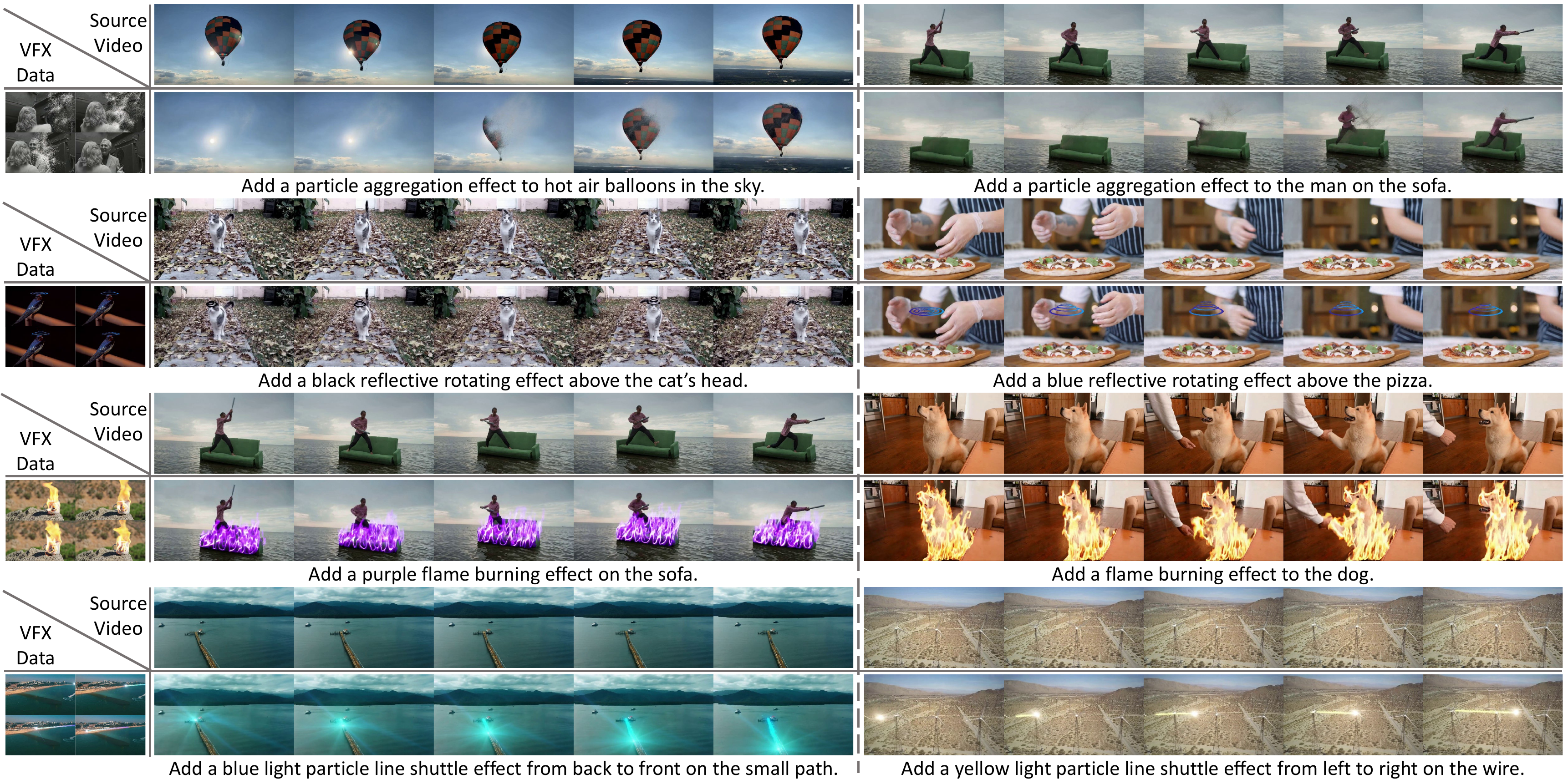}    
    \caption{
    \textbf{Additional Video VFX Editing Results of IC-Effect.} 
    }
    \label{fig:result_1}
\end{figure*}

\figref{result_0} and \figref{result_1} present additional video VFX editing results produced by IC-Effect. 
Across diverse effect categories, our IC-Effect accurately follows the textual prompts and generates visually coherent and well-integrated edits.

\subsection{Additional Qualitative Comparison of Video VFX Editing}

\begin{figure*}[!t]
    \centering
    \includegraphics[width=\linewidth]{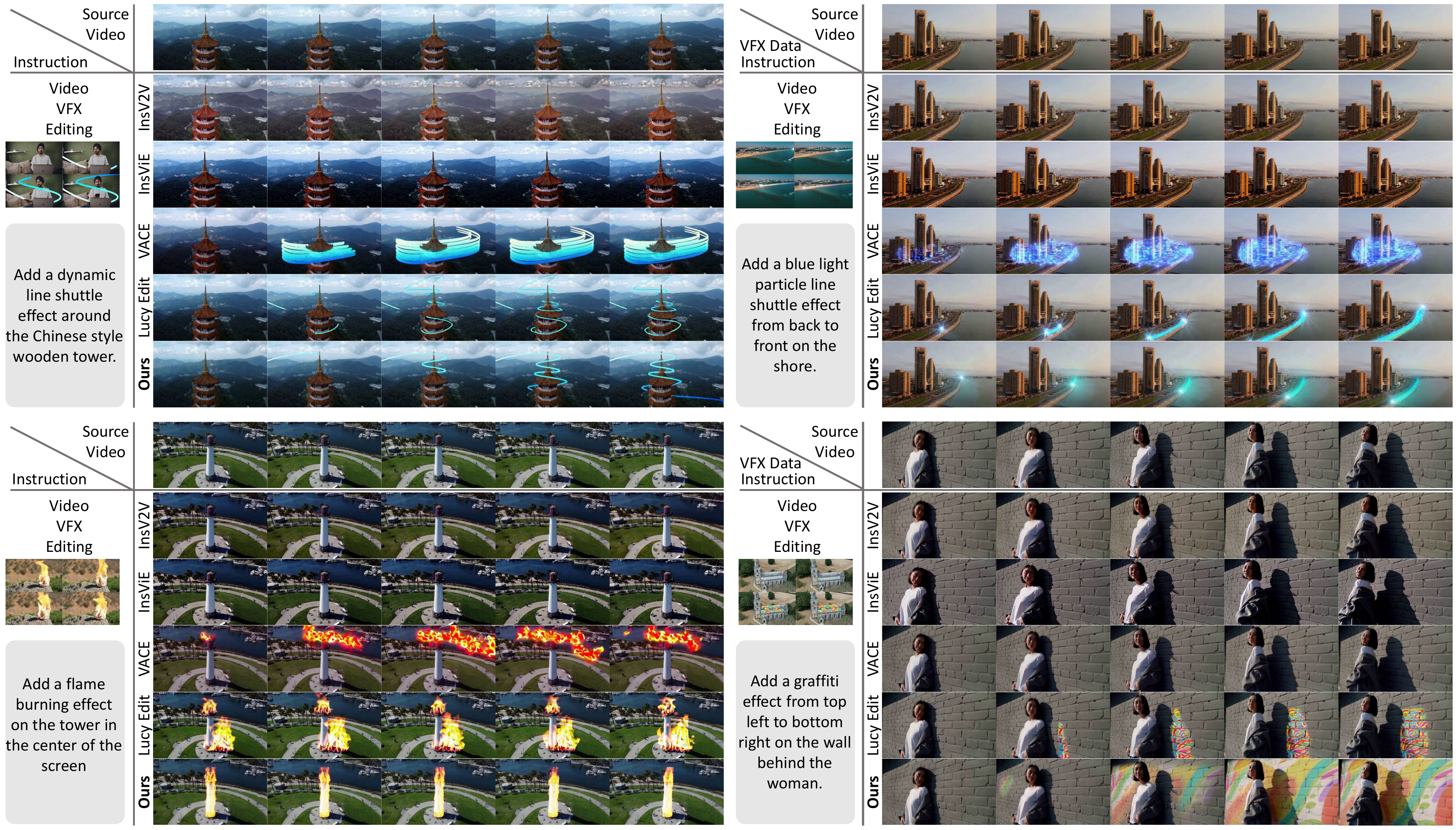}    
    \caption{
    \textbf{Additional Qualitative Comparison for Video VFX Editing.} 
    }
    \label{fig:vfx_0}
\end{figure*}
\begin{figure*}[!t]
    \centering
    \includegraphics[width=\linewidth]{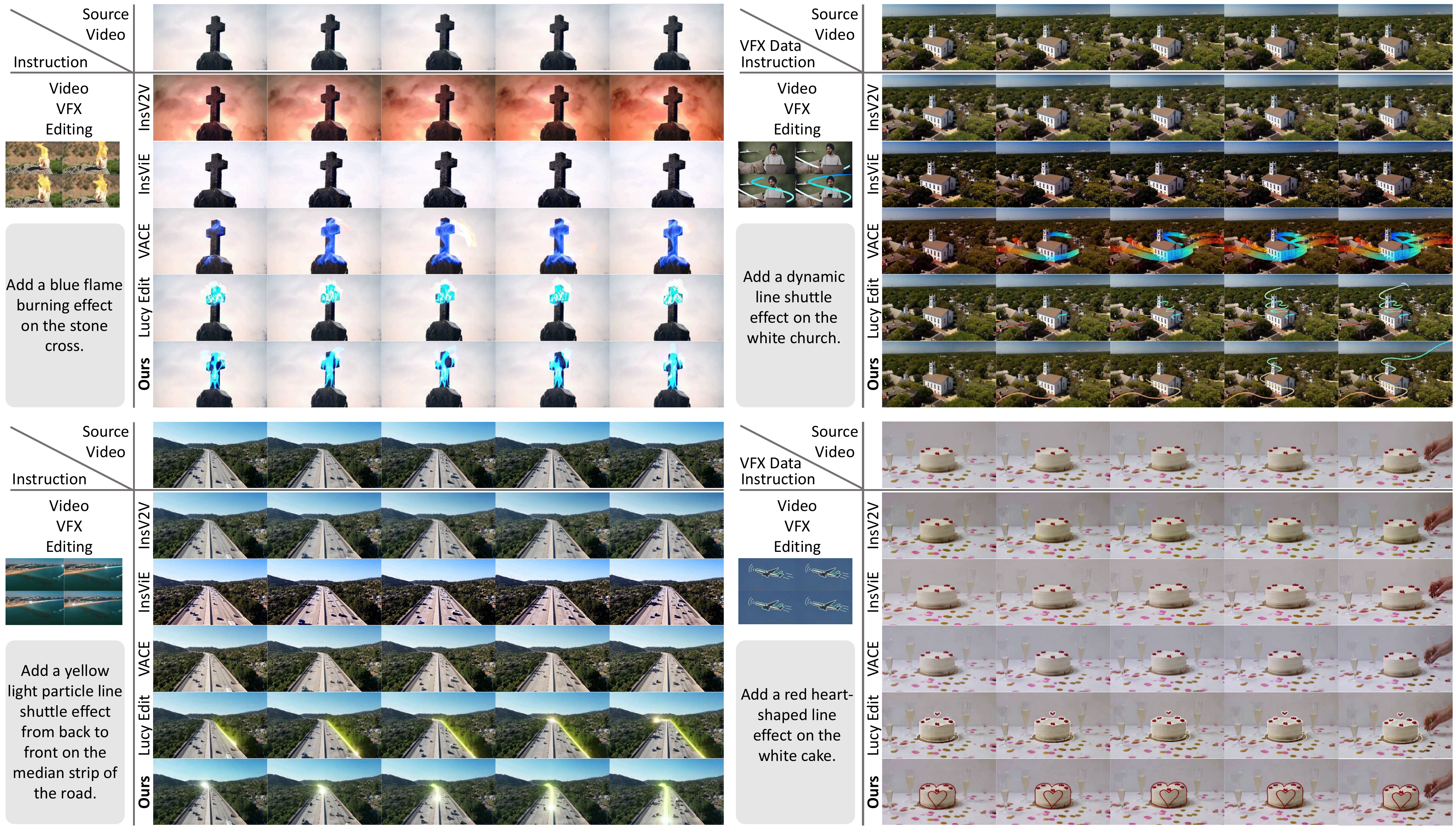}    
    \caption{
    \textbf{Additional Qualitative Comparison for Video VFX Editing.} 
    }
    \label{fig:vfx_1}
\end{figure*}
\begin{figure*}[!t]
    \centering
    \includegraphics[width=\linewidth]{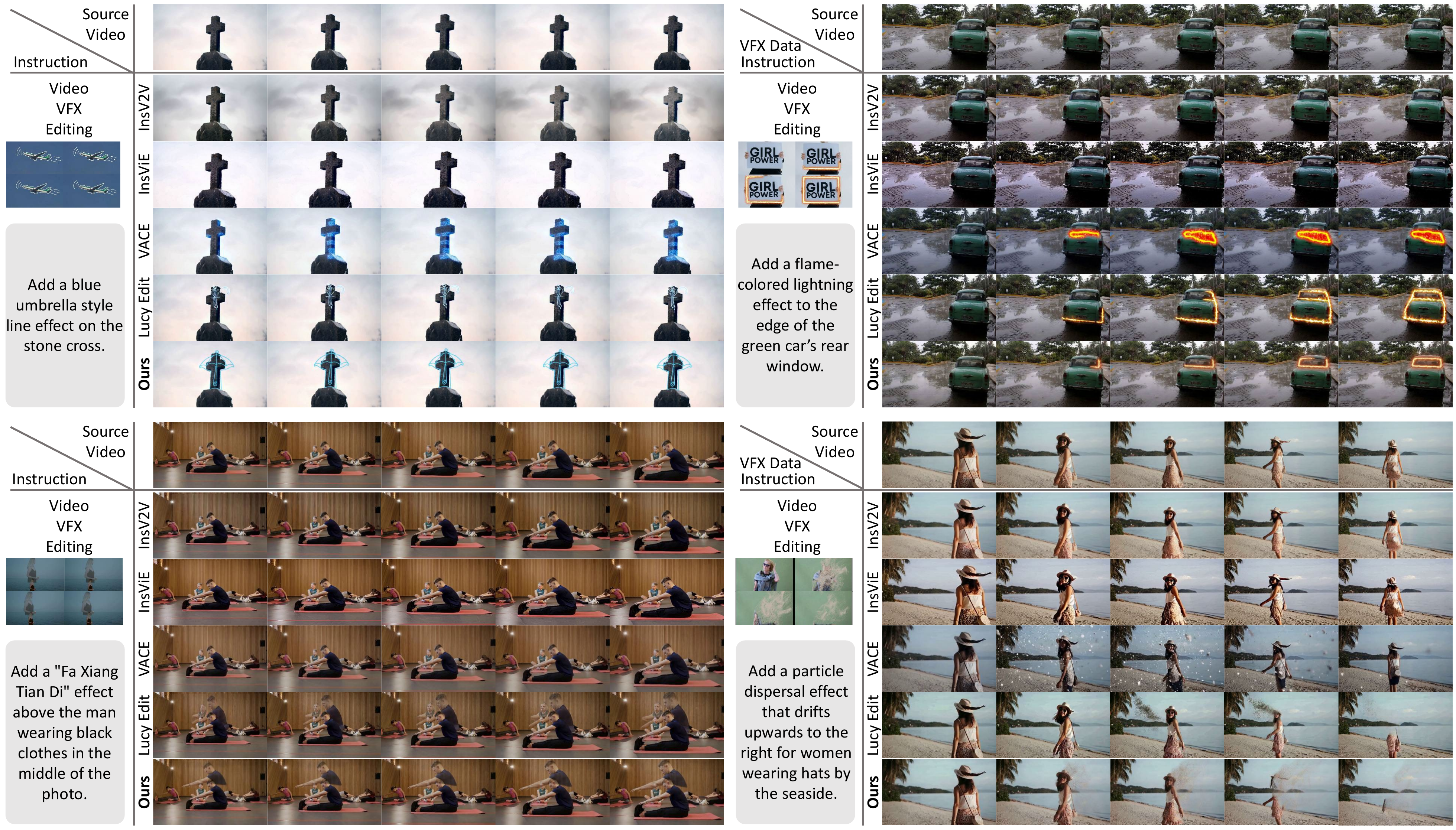}    
    \caption{
    \textbf{Additional Qualitative Comparison for Video VFX Editing.} 
    }
    \label{fig:vfx_2}
\end{figure*}

\figref{vfx_0} - \figref{vfx_2} present additional quantitative comparisons with baseline methods on video VFX editing. 
Compared with the baselines, IC-Effect consistently follows the textual prompts to edit the source video while preserving its spatial structure and temporal coherence.

\subsection{Additional Qualitative Comparison of Common Video Editing}

\begin{figure*}[!t]
    \centering
    \includegraphics[width=\linewidth]{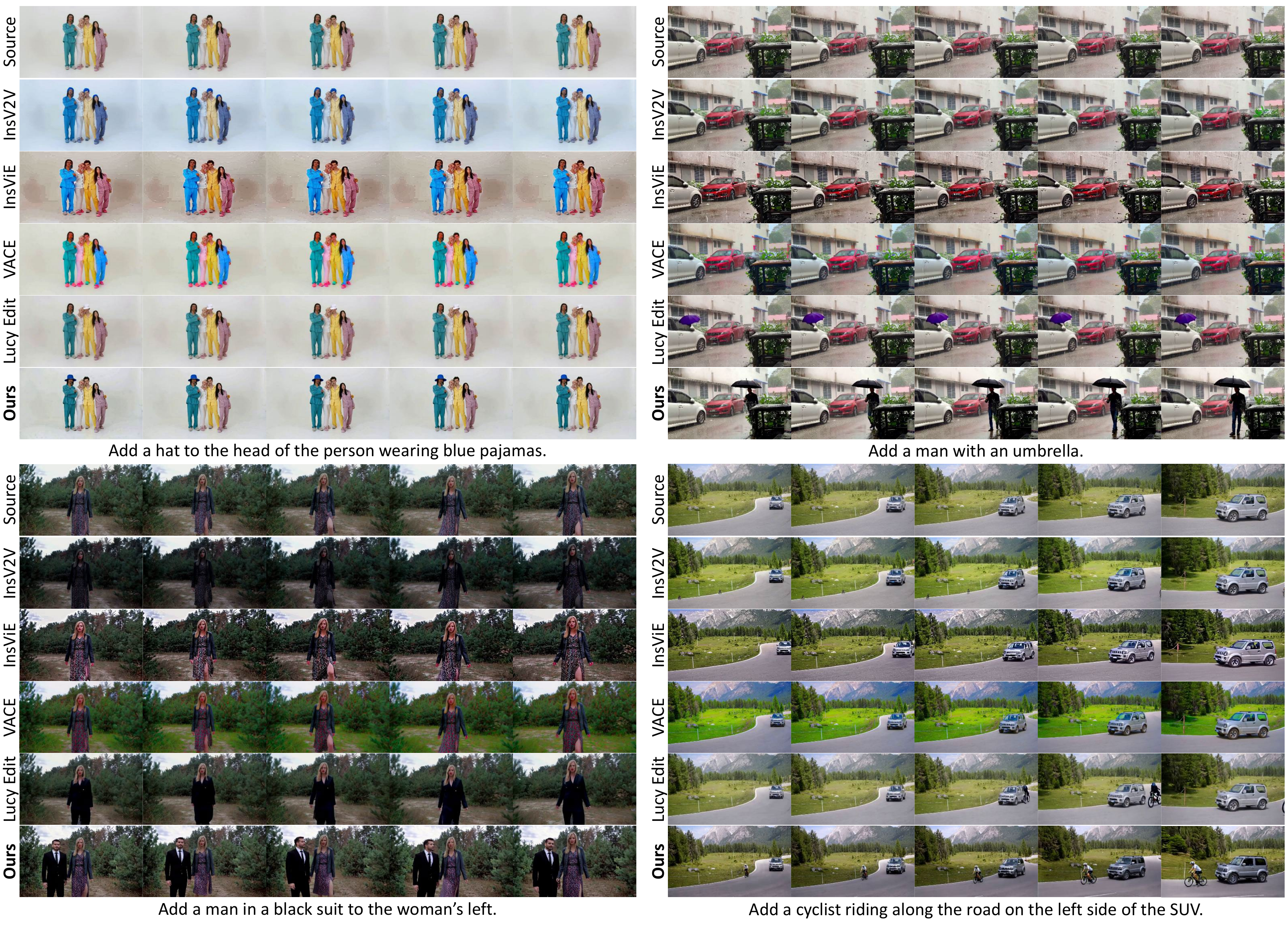}    
    \caption{
    \textbf{Additional Qualitative Comparison for Common Video Editing.} 
    }
    \label{fig:common_0}
\end{figure*}

\begin{figure*}[!t]
    \centering
    \includegraphics[width=\linewidth]{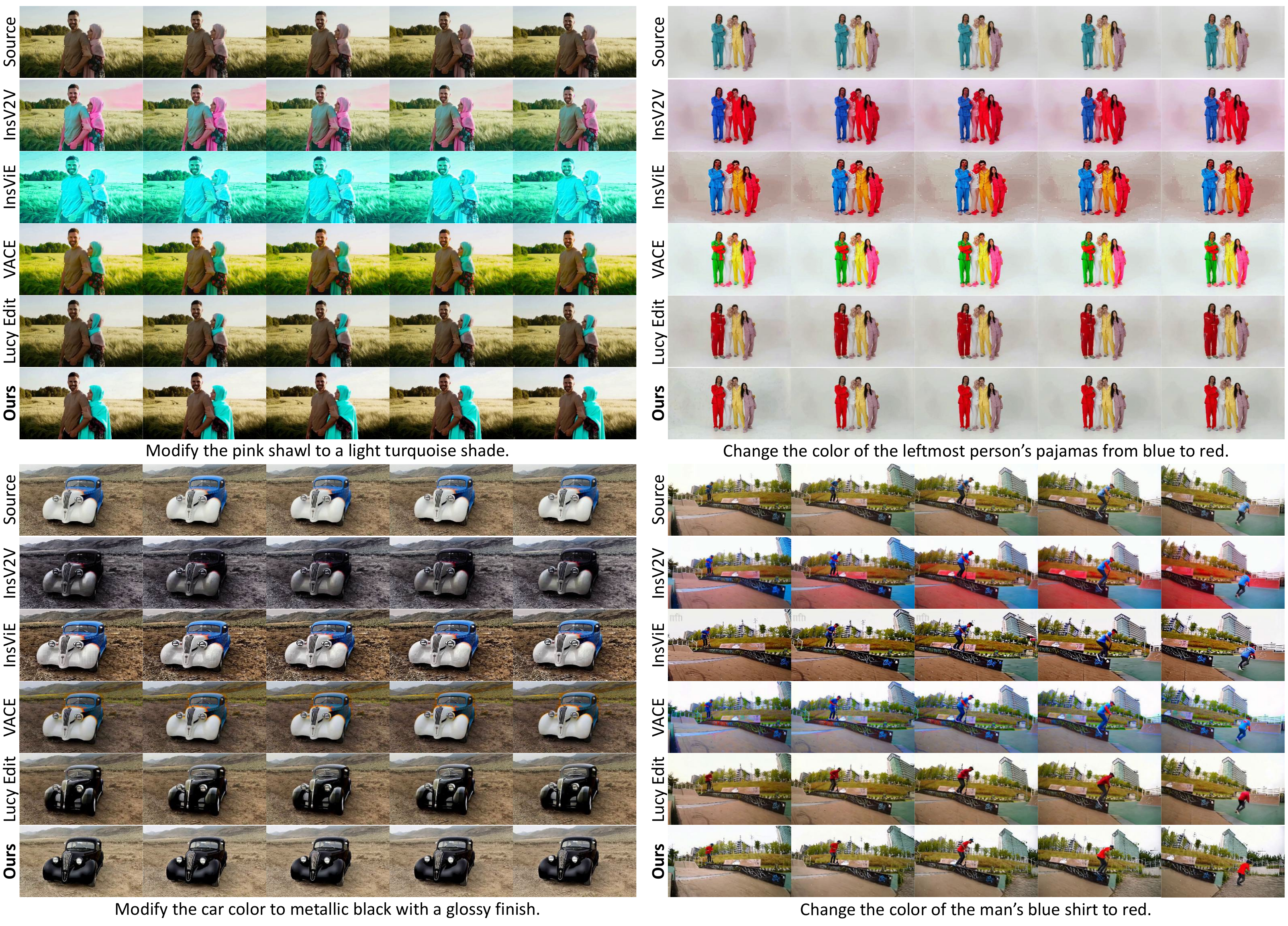}    
    \caption{
    \textbf{Additional Qualitative Comparison for Common Video Editing.} 
    }
    \label{fig:common_1}
\end{figure*}

\begin{figure*}[!t]
    \centering
    \includegraphics[width=\linewidth]{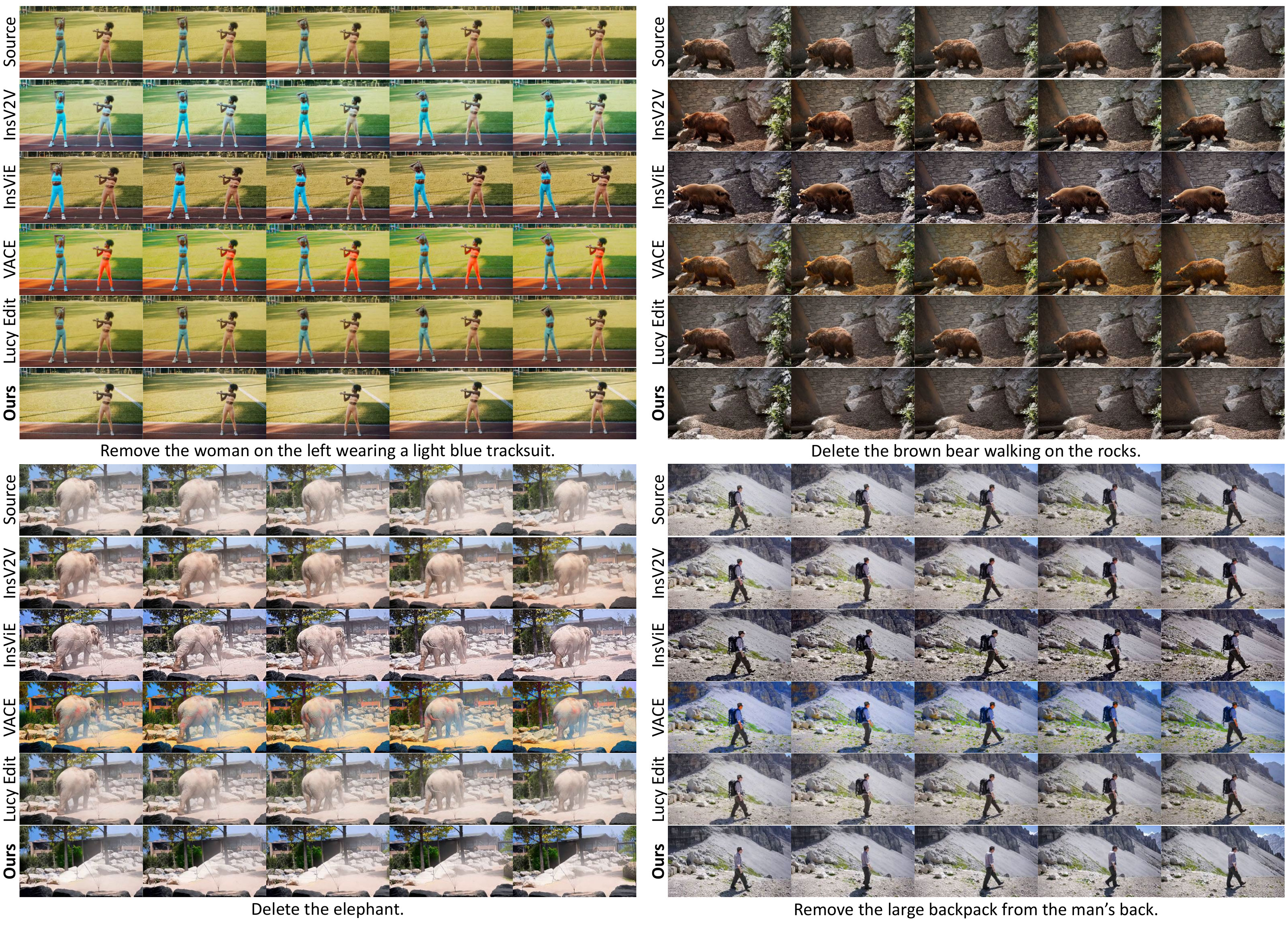}    
    \caption{
    \textbf{Additional Qualitative Comparison for Common Video Editing.} 
    }
    \label{fig:common_2}
\end{figure*}

\begin{figure*}[!t]
    \centering
    \includegraphics[width=\linewidth]{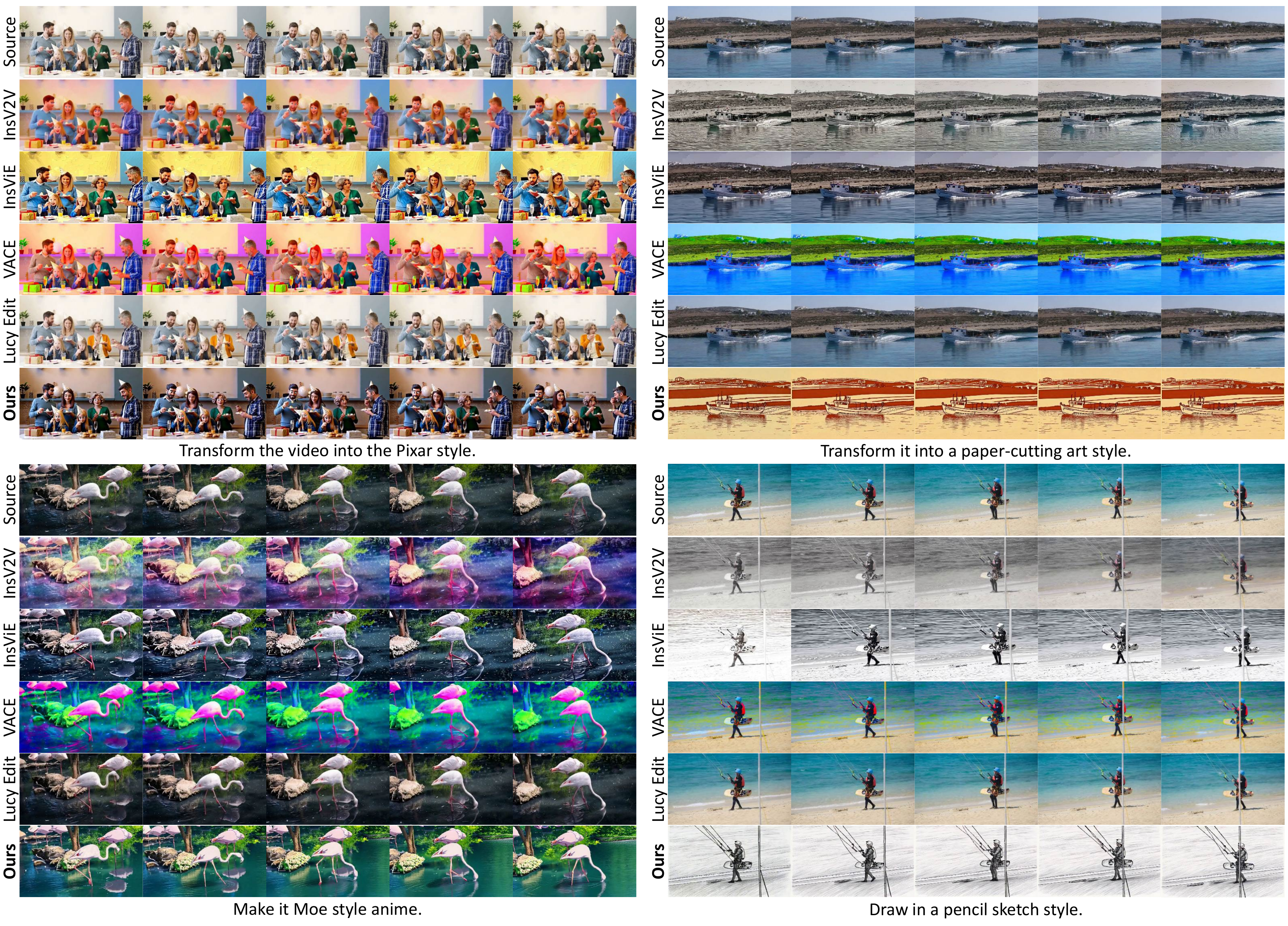}    
    \caption{
    \textbf{Additional Qualitative Comparison for Common Video Editing.} 
    }
    \label{fig:common_3}
\end{figure*}

\begin{figure*}[!t]
    \centering
    \includegraphics[width=\linewidth]{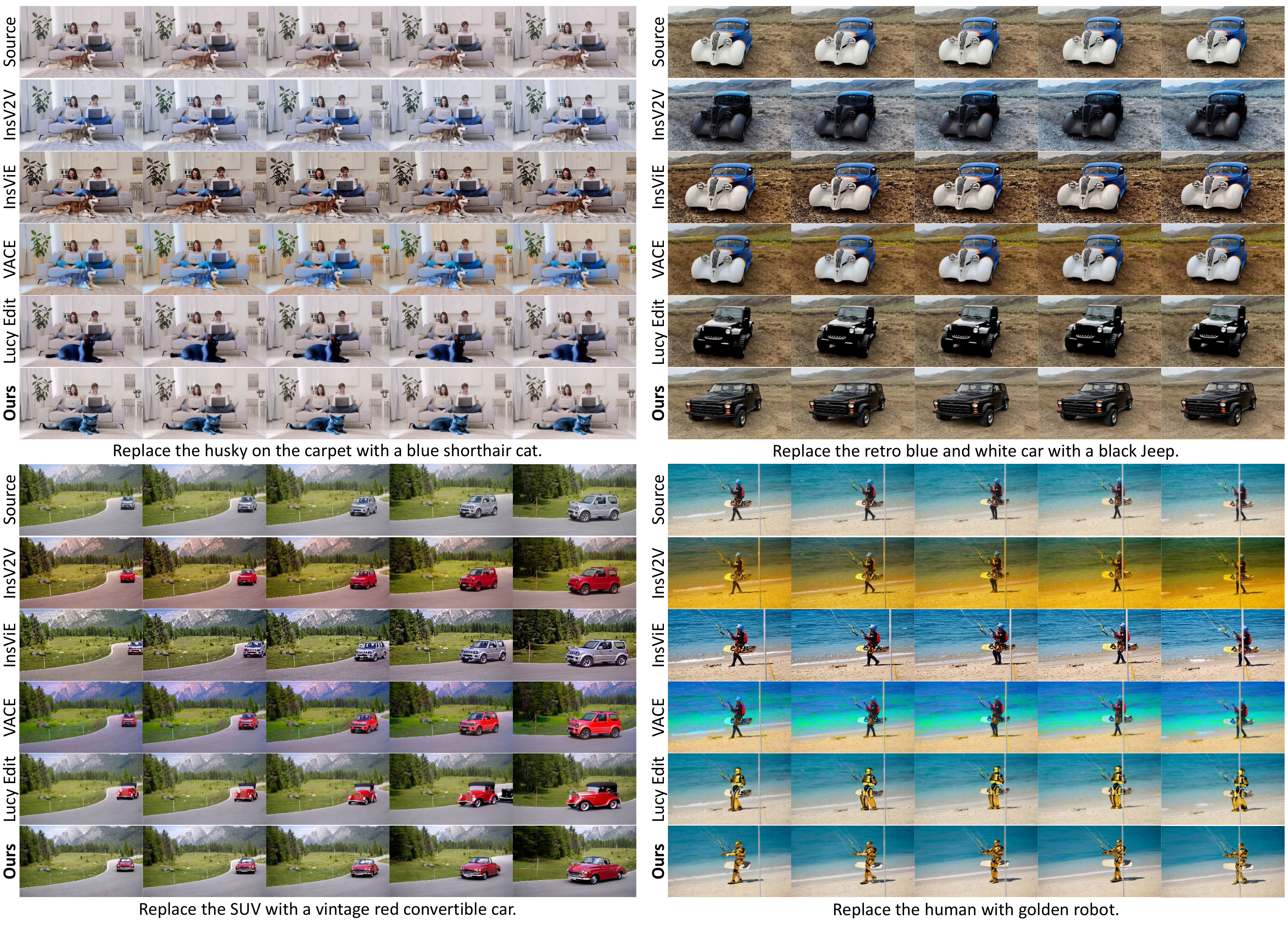}    
    \caption{
    \textbf{Additional Qualitative Comparison for Common Video Editing.} 
    }
    \label{fig:common_4}
\end{figure*}

\figref{common_0} - \figref{common_4} present qualitative comparisons between IC-Effect and existing methods across diverse common video editing tasks, including addition, removal, attribute modification, replacement, and style transfer.
Across these common editing scenarios, IC-Effect demonstrates superior performance: it accurately follows textual instructions, produces precise edits, and preserves the structural integrity of non-edited regions.

\end{document}